\documentclass[10pt,twocolumn,letterpaper]{article}
\PassOptionsToPackage{table}{xcolor}   
\usepackage[pagenumbers]{cvpr}   
\usepackage{times,graphicx,amsmath,amssymb,booktabs,capt-of,xcolor,pifont}
\usepackage{iftex}
\ifXeTeX
  \usepackage{fontspec}
\fi
\makeatletter
\renewcommand\paragraph{\@startsection{paragraph}{4}{\z@}{1.5ex \@plus .5ex \@minus .2ex}{-1em}{\normalfont\normalsize\bfseries}}
\makeatother
\AtEndPreamble{\crefname{appendix}{Appendix}{Appendices}\Crefname{appendix}{Appendix}{Appendices}}
\renewcommand{\floatpagefraction}{0.75}
\renewcommand{\dblfloatpagefraction}{0.75}
\usepackage[pagebackref,breaklinks,colorlinks]{hyperref}
\def\confName{arXiv}
\def\confYear{2026}
\newcommand{\bench}{EvOrderBench}
\newcommand{\PairHL}{\ensuremath{\mathrm{Pair}_{HL}}}   
\newcommand{\PairKP}{\ensuremath{\mathrm{Pair}_{KP}}}   
\definecolor{uzero}{HTML}{A3B1E0}\definecolor{uone}{HTML}{6175C4}\definecolor{utwo}{HTML}{2E3E8C}\definecolor{uheat}{HTML}{6175C4}\definecolor{tabink}{HTML}{243447}
\definecolor{markyes}{HTML}{2E7D4F}\definecolor{markpart}{HTML}{B07A1F}\definecolor{markno}{HTML}{B5413A}
\newcommand{\markY}{\textcolor{markyes}{\ding{51}}}\newcommand{\markP}{\textcolor{markpart}{$\blacktriangle$}}\newcommand{\markN}{\textcolor{markno}{\ding{55}}}
\newcommand{\uchip}[3]{{\setlength{\fboxsep}{1.3pt}\colorbox{#1}{\textcolor{#2}{#3}}}}
\title{\bench{}: When Do Sampled Frames\\Establish Event Order?}
\author{%
Xiaoqin Feng$^{1*}$\quad Yuqi Li$^{2*}$\quad Xinyuan Li$^{1}$\quad Neil Rodrigues$^{3}$\quad Xianglong Wang$^{3}$\\
Kuiye Ding$^{4}$\quad Yingli Tian$^{2\dagger}$\\[4pt]
{\normalsize $^1$University of Southern California\quad $^2$City University of New York\quad $^3$Wyze Inc.\quad $^4$University of Technology Sydney}\\[2pt]
{\small $^*$Equal contribution\quad $^\dagger$Corresponding author}\\[4pt]
{\small\url{https://xqfeng-josie.github.io/EvOrderBench/}}
}

\usepackage{placeins}
\usepackage[most]{tcolorbox}
\definecolor{ink}{HTML}{243447}\definecolor{chromehead}{HTML}{EFEDE7}\definecolor{chromeedge}{HTML}{D9D5CC}\definecolor{promptslot}{HTML}{9A5F12}
\newcommand{\slot}[1]{\textcolor{promptslot}{#1}}
\newtcolorbox{promptbox}[1]{enhanced, colback=white, colframe=chromeedge, colbacktitle=chromehead, coltitle=ink,
  title={#1}, fonttitle=\sffamily\bfseries\scriptsize, fontupper=\ttfamily\scriptsize\raggedright,
  fontlower=\sffamily\scriptsize\raggedright\color{ink!85}, boxrule=0.5pt, arc=2pt, left=4pt, right=4pt, top=3pt, bottom=3pt,
  toptitle=1.5pt, bottomtitle=1.5pt, segmentation style={dashed, draw=chromeedge}}
\newcommand{\promptnote}[1]{\tcblower #1}
\usepackage{etoolbox}
\definecolor{tblhead}{HTML}{DFDBF1}\definecolor{tblzebra}{HTML}{F3F1FA}\definecolor{tblrule}{HTML}{6E64A8}
\newcommand{\appendixtables}{%
  \setlength{\aboverulesep}{0pt}\setlength{\belowrulesep}{0pt}%
  \AtBeginEnvironment{tabular}{\renewcommand{\arraystretch}{1.22}\rowcolors{2}{white}{tblzebra}}%
  \let\origtoprule\toprule
  \renewcommand{\toprule}{\arrayrulecolor{tblrule}\origtoprule\rowcolor{tblhead}}}

\begin{document}
\maketitle
\begin{abstract}
Vision-language models (VLMs) are often asked which of two actions in a video happened first, given only a few sampled frames. Such frames can show that both actions occurred without showing which came first, so a reliable model should state an order only when the frames establish it. Forced-choice temporal benchmarks treat every input as answerable and reward lucky guesses, while answerability benchmarks do not separate frames that miss an action from frames that show both actions but not their order. We introduce \bench{}, a real-video benchmark of 579 event families. Each family fixes an episode and a question and varies only the frames: of two four-frame inputs, only one establishes the order, and two six-frame inputs extend the other one with either the frames that resolve the order or frames that do not. Every input is labeled for whether its frames establish each action and their order. Across sixteen open-weight configurations and three closed-source references, the best open models order nearly as accurately as the references when the frames establish the order, but withhold correctly on at most 83\% of the inputs where they do not, against at least 90\% for every reference; forced-choice accuracy cannot reveal this gap. Open models withhold mainly when the frames show neither action; once the frames show one or both, most of them commit to an order. Thinking narrows the gap, even when it is switched on within a single checkpoint, mainly by separating sufficient from insufficient evidence rather than by making models more cautious overall. Accurate ordering thus does not imply knowing when the frames fall short, and benchmarks should reward withholding alongside ordering.
\end{abstract}

\begin{figure*}[t]
\centering
\includegraphics[width=\textwidth]{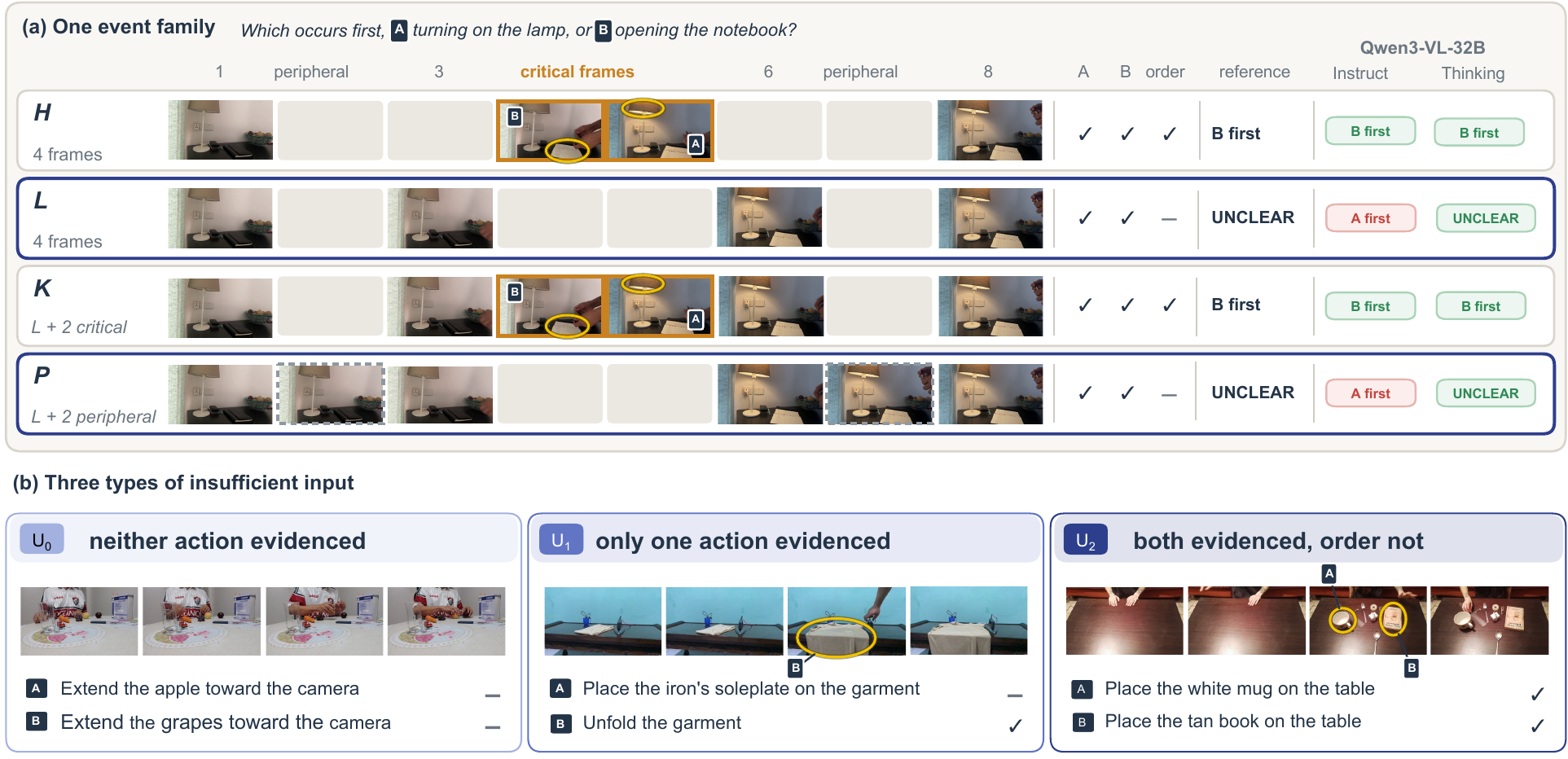}
\caption{\textbf{Same episode and question, different frames, different correct answers.} (a) Each row is one input of an event family, with its frames placed on the timeline of the episode. $H$ and $L$ have four frames each, but only $H$ includes the two critical frames (amber); $K$ adds them to $L$, and $P$ adds two peripheral frames instead (dashed). The frames of every input establish that both actions occurred (columns A and B), but only those of $H$ and $K$ establish their order (\checkmark/--: established or not), so the reference answer for $L$ and $P$ is UNCLEAR. Qwen3-VL-32B orders $H$ and $K$ correctly but commits on $L$ and $P$ to the opposite order; its Thinking variant withholds. (b) One real $L$ input for each type of insufficient evidence. Ovals and A/B tags are added for illustration and never shown to models; target descriptions are shortened.}
\label{fig:overview}
\end{figure*}
\section{Introduction}\label{sec:introduction}
Knowing which of two actions happened first is central to reviewing a recorded interaction, auditing a procedure, or monitoring task execution~\cite{coin,captaincook,robomonitor}. Vision-language models (VLMs) now answer such questions from sampled frames. When a few dozen frames are taken from a recording several minutes long, consecutive frames lie seconds apart, and a brief action may appear in one frame or in none. Its result often remains visible: a lamp that was off in one frame is lit in the next. The frames can then show that two actions occurred without showing which came first. In general, \emph{sampled frames establish the order of two actions only if they establish that each action occurred and how the two are related in time.} A reliable model should state the order exactly when this holds and withhold an answer otherwise.

\Cref{fig:overview}a shows such a case. The four frames of input $L$ show the lamp off and the notebook closed at the start and both changed at the end, but no frame in between shows which change came first. Adding the two frames in which the notebook is opened and the lamp turned on ($K$) settles the order, while adding two more views of the start and end states ($P$) does not. The episode and the question are the same for all of these inputs, but the correct answer is not. Qwen3-VL-32B answers $K$ correctly, yet on $L$ it states that the lamp was turned on first, the opposite of what $K$ shows. Its Thinking variant, which reasons before it answers, withholds on $L$ and $P$.

Current benchmarks cannot separate these behaviors. Temporal benchmarks assume that every question has an answer and score a forced choice among orders or temporal attributes~\cite{tempcompass,tomato,vector}, and TOMATO's evaluation prompt even tells models to guess when the frames are insufficient~\cite{tomato}. A lucky guess on $L$ then counts as temporal understanding, and a model that rightly withholds gains nothing. Answerability benchmarks let models abstain and label whether a question can be answered from the input~\cite{uvqa,virtuebench,stormbench}, some also with a coarse cause such as occlusion~\cite{stormbench,trapsbench}, but they do not separate a missing action from an unresolved order between two observed actions, and in VirtueBench answerability changes with the frame count~\cite{virtuebench}. A refusal may thus reflect a missing action, an unresolved order, or a shorter input (\cref{tab:compare}).

To separate these behaviors, a benchmark needs three things: labels that state what each input establishes, for each action and for their order; matched inputs that change this evidence while the question and the number of frames stay fixed; and scoring that rewards ordering and withholding together. We introduce \bench{}, which provides all three. It contains 579 \emph{event families} from three real-video datasets of hand--object interaction. Each family fixes a recorded episode and a question about two target actions and varies only the frames (\cref{fig:overview}a): two four-frame inputs, one that includes the frames resolving the order and one that omits them, and two six-frame inputs that extend the latter with either those resolving frames or two frames that leave the order open. All 2,316 inputs are labeled, under explicit criteria and with human review, for whether the frames establish each action and their order. An input whose frames do not establish the order is \emph{insufficient}, and its labels also say what is missing: evidence for both actions, for one of them, or only for their order. Every input is queried under both A/B label assignments, and joint metrics require related inputs of a family to be answered correctly together.

We evaluate sixteen open-weight configurations from nine model series, covering scale pairs, supervised and RL post-training, and Thinking, with two Gemini models and one GPT model as closed-source references. Three findings emerge. First, ordering and withholding come apart. When the frames establish the order, the best open models nearly match the references, yet only two open configurations withhold correctly on more than 70\% of insufficient inputs (at most 83\%), against at least 90\% for every reference. A forced-choice score cannot see this gap and would even credit the unsupported answers, which match the true order more often than chance. Second, withholding depends on what is missing. Open models withhold mainly when the frames show neither action; once the frames show at least one action, most commit to an order, whether the second action or only their order is missing. A matched diagnostic subset, which changes the missing evidence within a family, shows the same. Third, Thinking narrows the gap, including when it is switched on within one checkpoint, mainly by separating sufficient from insufficient evidence rather than by making models more cautious overall. Yet even the best open configuration answers all eight queries of a family correctly for only two thirds of the families. Most open VLMs, in short, treat evidence for either action as a license to answer.

Our contributions are threefold. (1) We formulate event ordering as a judgment conditioned on the frames a model receives, and separate two ways in which the evidence can fall short: a missing action and an unresolved order between two observed actions. (2) We introduce \bench{}: 579 real-video event families with matched inputs, occurrence and order labels for every input, and metrics that score ordering and withholding jointly. (3) We evaluate sixteen open-weight configurations and three closed-source references, and show that the gap between them is widest in withholding, above all once the frames show at least one action, and that Thinking narrows it by separating sufficient from insufficient evidence. The benchmark, labels, prompts, and evaluation code are available at \url{https://xqfeng-josie.github.io/EvOrderBench/}.

\section{Related Work}
\begin{table}[t]
\centering\footnotesize\setlength{\tabcolsep}{2.5pt}
\begin{tabular*}{\linewidth}{@{\extracolsep{\fill}}lcccc@{}}
\toprule
Benchmark & \shortstack{Real\\video} & \shortstack{Model\\withholds} & \shortstack{Same-size\\matched inputs} & \shortstack{Action vs.\\order} \\
\midrule
TempCompass~\cite{tempcompass} & \markP & \markN & \markP & \markN \\
TOMATO~\cite{tomato} & \markP & \markN & \markP & \markN \\
UVQA~\cite{uvqa} & \markY & \markY & \markN & \markP \\
VirtueBench~\cite{virtuebench} & \markY & \markY & \markP & \markN \\
RevisitQA~\cite{stormbench} & \markY & \markY & \markP & \markP \\
TRAPSBench~\cite{trapsbench} & \markN & \markY & \markP & \markP \\
Caved or Convinced~\cite{cavedconvinced} & \markP & \markP & \markY & \markN \\
\midrule
\bench{} (ours) & \markY & \markY & \markY & \markY \\
\bottomrule
\end{tabular*}
\caption{\textbf{Comparison with related benchmarks.} Same-size matched inputs differ in evidence, not frame count; action vs.\ order: labels tell a missing action from an unresolved order. \markP: partial (\eg, edited clips, unequal lengths, external abstention test, coarse causes).}
\label{tab:compare}
\end{table}

\paragraph{Temporal video understanding.}
Video benchmarks span object interactions, causal and compositional reasoning, broad task suites, and long contexts~\cite{something,nextqa,star,agqa,mvbench,videomme,egoschema,longvideobench,mlvu}. Several evaluations target temporal order directly. Test of Time finds that video-language models struggle with before/after relations~\cite{testoftime}, TempCompass and TVBench remove cues that let models answer without temporal reasoning~\cite{tempcompass,tvbench}, TOMATO measures how much models gain from multiple frames and from their order~\cite{tomato}, and VECTOR, Vinoground, and TimeBlind probe event order and temporal composition with synthetic or minimally different videos~\cite{vector,vinoground,timeblind}. All of these assume that the input determines one correct answer. \bench{} asks a prior question: whether the input establishes the order at all.

\paragraph{Answerability and abstention.}
A model should withhold an answer that its input does not support~\cite{squad2,reliablevqa}, yet evaluations that score only correct answers reward guessing instead~\cite{videoevalpro}, which has been argued to sustain hallucination~\cite{whyhallucinate}. Multimodal and language benchmarks show that models still answer when the input supports no answer~\cite{mmupd,abstentionbench}, and reasoning fine-tuning can even make language models abstain less~\cite{abstentionbench}. In video, UVQA trains models to refuse questions that do not match the video, with absent objects, undepicted relations, and mismatched attributes as separate causes~\cite{uvqa}, and absent-answer detection removes the correct option~\cite{absentanswer}.

Closest to our setting, VirtueBench labels answerability at each frame-sampling level of long videos~\cite{virtuebench}, RevisitQA labels each query, at the time it is asked, as known or uncertain and records the cause~\cite{stormbench}, and TRAPSBench pairs answerable physics simulations with occluded or truncated versions~\cite{trapsbench}. For event order, Caved or Convinced shows on composited videos that sparse sampling can leave a true and a false order claim indistinguishable, and lets a forward--reverse frame test, rather than the model, decide when to abstain~\cite{cavedconvinced}. Confidence-based abstention has also been compared on the same video questions under full and truncated frame sets, which differ in length~\cite{abstentionknobs}. We share the premise that an answer must be supported by the given input. Unlike these benchmarks, \bench{} combines real videos, matched inputs that change the evidence at a fixed frame count, and labels that separate a missing action from an unresolved order (\cref{tab:compare}).

\paragraph{Hallucination and matched evaluation.}
Hallucination benchmarks test claims that the visual input does not support~\cite{pope,hallusionbench,videohallucer,vidhal,mohallbench,noah}. Ordering an action that the frames do not show is a temporal claim of this kind, and our inputs that show only one of the two actions isolate it. Matched designs expose failures that marginal accuracy hides, in images~\cite{winoground,sugarcrepe,visualanswerability} and in videos~\cite{mvp,twinpast}. We apply matched, jointly scored inputs to the evidence for event order.

\section{\bench{}}\label{sec:benchmark}
\bench{} tests whether an event-order judgment is supported by the frames a model receives. Its unit is an \emph{event family}: one recorded episode, a question about two target actions, and four observation sets ($H$, $L$, $K$, and $P$), each a selection of frames that forms one input to a model (\cref{fig:overview}). \Cref{sec:task_definition,sec:families,sec:metrics} provide the labels, matched inputs, and joint scoring that \cref{sec:introduction} calls for.

\subsection{Task Definition}\label{sec:task_definition}
An event family $e$ fixes a question $q_e=(A_e,B_e,s_e)$: two target descriptions and an optional scope statement $s_e$ (\eg, ``the first time''). For each observation set $c$, the input $X_{e,c}$ is a chronologically ordered set of frames, and the model answers
\[
\hat{y}_{e,c}\in\{A_e\prec B_e,\ B_e\prec A_e,\ \mathrm{UNCLEAR}\},
\]
judging which action occurs first in the recorded segment \emph{as established by $X_{e,c}$}. Each input carries a reference label $y_{e,c}$ from the same set. The question does not presuppose that either action occurs. Evidence from the family's other observation sets or from unsampled source frames cannot justify an answer, so inputs from the same episode may have different correct answers.

\paragraph{Occurrence and order labels.}
Each target description fixes what counts as its action: a specified result, a visible operation, or a contact (\cref{app:annotation}). For each target $t\in\{A,B\}$, the occurrence label $o_{e,c,t}$ is ESTABLISHED when the frames show the described action, or a state change that only this action can produce (\eg, the lamp off in one frame and lit in a later one), and NOT\_ESTABLISHED otherwise. The latter is an evidential judgment, not a claim that the action did not happen. Object presence or nearby hand motion does not establish an action, and neither does an end state without the state before it: the question concerns what happens within the segment, and the state may have existed before the first frame. Assuming that states persist between frames unless a frame shows otherwise, an established occurrence is located at the frames that show it, or between the two consecutive frames that bracket its state change.

A definite order requires that both occurrences are established and that one lies entirely before the other. When both changes fall between the same two frames, as in $L$ of \cref{fig:overview}a, the order is UNCLEAR even if the end states make one order more plausible (\eg, one object left on top of the other), because the answer must be checkable against the given frames. An ambiguous target, recipient, or event boundary makes the question itself ill-posed and leads to revision or exclusion, not to an UNCLEAR label.

\paragraph{Evidence types.}
An action is \emph{evidenced} in an input when its occurrence label is ESTABLISHED; an input is \emph{orderable} when its reference label is an order and \emph{insufficient} when it is UNCLEAR; and a model \emph{withholds} when it answers UNCLEAR. Let $n_{e,c}=|\{t:o_{e,c,t}=\mathrm{ESTABLISHED}\}|$ count the evidenced targets; each insufficient input belongs to exactly one type $U_k=\{(e,c): y_{e,c}=\mathrm{UNCLEAR},\ n_{e,c}=k\}$ with $k\in\{0,1,2\}$ (\cref{fig:overview}b). $U_0$ and $U_1$ miss at least one action, which frames showing it would supply; $U_2$ misses only the order, which a frame showing one action complete before the other occurs would supply.

\subsection{Event Families}\label{sec:families}
Each family contains two four-frame inputs, $H$ and $L$, and two six-frame inputs, $K$ and $P$:
\[
H\cup L=K,\qquad L\subset P,\qquad |K\setminus L|=|P\setminus L|=2.
\]
$H$ (high evidence) includes the intermediate frames that resolve the order, and $L$ (low evidence) omits them, usually taking two frames near each end of the segment and none in between; all four inputs share the same first and last frame, a median of 17\,s apart. $K$ adds to $L$ the two \emph{critical} frames that distinguish $H$ from $L$, and $P$ adds two \emph{peripheral} frames, one near each end of $L$, that are not meant to resolve the order. The frames are selected rather than sampled at a fixed rate, so that the evidence can change while the episode, the question, and the frame count stay fixed.

The design provides two controls. In the equal-count control, $H$ vs.\ $L$ and $K$ vs.\ $P$ compare inputs with the same number of frames, so differences in what the frames establish, and in model behavior, cannot be attributed to input length. They differ in coverage instead: $L$ and $P$ leave unsampled the interval in which the actions take place (\cref{fig:timing}). Consecutive frames of $H$ lie a median of 5.1\,s apart, whereas $L$ places two frames near each end of the window in 545 of the 549 families with timestamps, and consecutive frames of $P$ lie a median of 0.5\,s apart, because the frames it adds sit next to frames already in $L$; the critical frames that $K$ adds fall inside the unsampled gap. In the evidence-addition control, the label of $K$ provides the reference order against which we check whether unsupported answers on $L$ and $P$ are informed guesses, the $L\!\to\!K$ comparison tests whether models order correctly once the missing frames are included, and $L\!\to\!P$ whether frames without that evidence leave their answers unchanged.

\subsection{Data Collection and Annotation}
Episodes come from three public datasets of real hand--object interaction: Perception Test~\cite{perceptiontest} (532 families), HO-Cap~\cite{hocap} (30), and CaptainCook4D~\cite{captaincook} (17). Candidates were screened for target pairs with distinguishable objects, visible interactions, and interpretable boundaries.

Candidate questions, frame sets, and provisional labels were drafted with Claude, which is not among the evaluated models. At least two human annotators then checked every item against the criteria of \cref{sec:task_definition} and could revise its wording or scope, replace frames, correct its labels, or exclude it; revised items returned for review, and every released label was accepted by all annotators. Annotators excluded 18 of the 597 reviewed families (3.0\%), all for poor video quality (\cref{app:review}). Each review batch contained a single observation set, such as only $L$ inputs, so labels could not draw on the other inputs of a family. On 32 randomly drawn families, two annotators labeling independently and without the provisional labels agreed on all occurrence and order labels for 125 of the 128 inputs (97.7\%).

\begin{figure}[t]
\centering
\includegraphics[width=\linewidth]{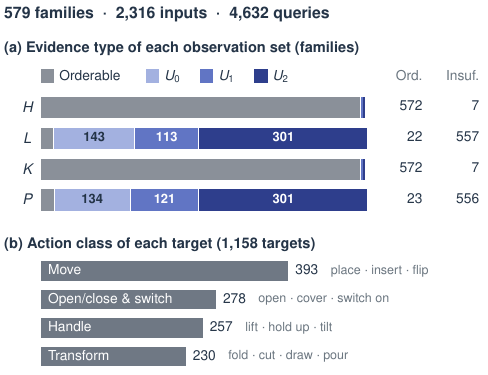}
\caption{\textbf{$H$ and $K$ are nearly always orderable; $L$ and $P$ span all three evidence types.} (a) Evidence type of each observation set, in families. (b) Targets grouped by what the action changes, with representative actions in gray.}
\label{fig:benchmark_stats}
\end{figure}

\subsection{Dataset Statistics}\label{sec:stats}
\bench{} contains 579 families (\cref{fig:benchmark_stats}). Their 1,158 targets fall into four action classes defined by what the action changes: move, open/close \& switch, handle, and transform (\cref{fig:benchmark_stats}b). Of the 2,316 inputs, 1,189 are orderable and 1,127 insufficient: 277 in $U_0$, 240 in $U_1$, and 610 in $U_2$ (\cref{fig:benchmark_stats}a). In 549 families, $H$ and $K$ are orderable while $L$ and $P$ are insufficient; the remaining 30, 22 of them orderable throughout, keep whatever labels their frames support, so labels follow the frames rather than an input's role. Apart from these 22, the insufficient inputs of a family miss either an action ($U_0$ or $U_1$; 256 families) or only the order ($U_2$; 301); no family mixes the two. Evidence type also depends on the kind of action: handle actions such as lifting or touching rarely leave a lasting change, so 87\% of handle targets are not evidenced in $L$, against 14--25\% in other classes. Comparisons between types therefore also compare different events and actions, which \cref{sec:exp_analysis} controls for with a matched diagnostic subset.

\subsection{Evaluation Protocol and Metrics}\label{sec:metrics}
Each input is queried independently under the original and the swapped assignment of the answer letters A and B to the two targets, giving 4,632 queries per model; swapping reverses the correct order and preserves UNCLEAR, which cancels any fixed preference for one answer letter. We report overall accuracy on all queries, accuracy on orderable inputs (\emph{Ord.}), accuracy on insufficient inputs (\emph{Insuf.}), and the correct-UNCLEAR rate within each of $U_0$, $U_1$, and $U_2$.

Marginal accuracies can be inflated by a fixed policy: always answering UNCLEAR scores 100\% on Insuf. Joint metrics therefore require a family's related inputs to be answered correctly together. With $C_{e,c,a}\in\{0,1\}$ the correctness of family $e$, observation set $c$, and assignment $a$,
\[
\mathrm{Pair}_{HL}=\frac{1}{2|\mathcal{E}_{HL}|}\sum_{e\in\mathcal{E}_{HL}}\sum_{a=1}^{2}C_{e,H,a}\,C_{e,L,a},
\]
where $\mathcal{E}_{HL}$ contains the 550 families with orderable $H$ and insufficient $L$; a model must order $H$ and withhold on $L$. \PairKP{} is defined analogously on $K$ and $P$ (549 families), and Family requires all eight queries of a family to be correct (579 families).

To test whether models recognize missing evidence, a diagnostic occurrence query, posed separately to six instruction models and two Thinking configurations, asks whether each target is ESTABLISHED in the same input. Confidence intervals use a cluster bootstrap with 5,000 resamples over the 558 recording and participant groups identified in the source metadata, keeping correlated families together.

\begin{figure}[t]
\centering\footnotesize
\begin{promptbox}{Order query (main prompt, every model)}
Use only the supplied images as evidence. When there are multiple images, their presentation order is chronological, but they are sampled frames, not a complete recording.\par The question or statement does NOT guarantee that either action occurred. Do not assume that an unseen action happened earlier or will happen later. Seeing an object, a hand near it, or one action alone does not by itself establish the order of two actions. A static state can be used only when it actually establishes the relevant events and their order. Do not infer order from typical routines or object position alone.\par\smallskip Allowed answer labels: A, B, UNCLEAR. Choose A or B only when the supplied images establish the order of both specif{}ied actions. Otherwise choose UNCLEAR.\par Question: Using only the supplied images, which occurs f{}irst in the recorded segment: \slot{Turn on the table lamp with the beige shade.} (A), or \slot{Open the black-covered notebook.} (B)?\par\smallskip Return exactly one of the allowed answer labels above. Do not output an explanation, JSON, Markdown, or any other text.
\promptnote{Input $L$ of \cref{fig:overview}, original assignment. Only the colored target descriptions change between families; 16 families add a reviewed scope sentence. The swapped query exchanges the two descriptions (\cref{fig:prompts_main}).}
\end{promptbox}
\caption{\textbf{The prompt states what an answer requires}, verbatim. The frames precede the text as separate images in the model's native chat template; no system prompt is added.}
\label{fig:prompt}
\end{figure}

\begin{table*}[t]
\centering\footnotesize
\begin{tabular*}{\textwidth}{@{\extracolsep{\fill}}lrrrrrrrrr@{}}
\toprule
 & \multicolumn{3}{c}{Per query} & \multicolumn{3}{c}{Correct UNCLEAR by type} & \multicolumn{3}{c}{Joint} \\
\cmidrule(lr){2-4}\cmidrule(lr){5-7}\cmidrule(lr){8-10}
Model & Overall & Ord. & Insuf. & \uchip{uzero}{tabink}{$U_0$} & \uchip{uone}{white}{$U_1$} & \uchip{utwo}{white}{$U_2$} & \PairHL & \PairKP & Family {\scriptsize(all 8)} \\
\midrule
\textit{Always A} & 25.7 & 50.0 & 0.0 & 0.0 & 0.0 & 0.0 & 0.0 & 0.0 & 0.0 \\
\textit{Always UNCLEAR} & 48.7 & 0.0 & 100.0 & 100.0 & 100.0 & 100.0 & 0.0 & 0.0 & 1.2 \\
\midrule
Gemma 4 12B & 62.3 & 56.6 & 68.3 & \cellcolor{uheat!35}86.8 & \cellcolor{uheat!26}64.8 & \cellcolor{uheat!24}61.2 & 33.5 & 36.3 & 20.0 \\
GLM-4.6V-Flash & 33.1 & 51.9 & 13.2 & \cellcolor{uheat!19}48.2 & \cellcolor{uheat!2}5.6 & \cellcolor{uheat!0}0.2 & 8.0 & 6.7 & 0.3 \\
InternVL3.5-8B & 43.7 & 73.5 & 12.2 & \cellcolor{uheat!14}36.1 & \cellcolor{uheat!4}9.0 & \cellcolor{uheat!1}2.7 & 8.7 & 11.0 & 7.3 \\
InternVL3.5-14B & 44.5 & 76.3 & 11.0 & \cellcolor{uheat!13}31.9 & \cellcolor{uheat!4}10.4 & \cellcolor{uheat!1}1.6 & 9.5 & 9.1 & 7.3 \\
Keye-VL-1.5-8B & 46.8 & 51.6 & 41.9 & \cellcolor{uheat!24}58.8 & \cellcolor{uheat!11}27.5 & \cellcolor{uheat!16}39.8 & 16.8 & 15.9 & 5.0 \\
LLaVA-OneVision-2-8B & 42.2 & 81.3 & 0.9 & \cellcolor{uheat!2}3.8 & \cellcolor{uheat!0}0.0 & \cellcolor{uheat!0}0.0 & 1.1 & 0.8 & 3.1 \\
MiMo-VL-7B-SFT & 44.5 & 61.4 & 26.6 & \cellcolor{uheat!24}60.1 & \cellcolor{uheat!12}30.0 & \cellcolor{uheat!4}10.1 & 19.1 & 15.8 & 5.9 \\
MiMo-VL-7B-RL & 54.5 & 67.3 & 40.9 & \cellcolor{uheat!30}73.8 & \cellcolor{uheat!17}42.9 & \cellcolor{uheat!10}25.2 & 28.1 & 27.0 & 10.7 \\
MiniCPM-V4.7 & 58.7 & \underline{93.9} & 21.6 & \cellcolor{uheat!20}50.0 & \cellcolor{uheat!1}1.5 & \cellcolor{uheat!7}16.6 & 21.3 & 21.5 & 19.3 \\
Qwen3-VL-8B & 54.4 & 80.9 & 26.6 & \cellcolor{uheat!27}66.6 & \cellcolor{uheat!9}22.7 & \cellcolor{uheat!4}9.9 & 23.7 & 21.1 & 16.8 \\
Qwen3-VL-32B & 61.5 & 88.6 & 32.9 & \cellcolor{uheat!30}74.4 & \cellcolor{uheat!3}6.5 & \cellcolor{uheat!10}24.4 & 29.0 & 30.7 & 23.1 \\
Qwen3.8-27B & 73.4 & 77.0 & 69.5 & \cellcolor{uheat!36}\underline{91.0} & \cellcolor{uheat!20}50.8 & \cellcolor{uheat!27}67.1 & 52.2 & 52.6 & 33.0 \\
\midrule
MiMo-VL-7B-RL (Thinking) & \underline{84.2} & 86.6 & \underline{81.7} & \cellcolor{uheat!36}89.7 & \cellcolor{uheat!29}\underline{73.1} & \cellcolor{uheat!33}\textbf{81.4} & \underline{69.9} & \underline{72.0} & \underline{51.5} \\
Qwen3-VL-8B-Thinking & 72.6 & 86.7 & 57.8 & \cellcolor{uheat!27}68.4 & \cellcolor{uheat!20}49.8 & \cellcolor{uheat!22}56.1 & 51.3 & 50.3 & 32.1 \\
Qwen3-VL-32B-Thinking & 81.2 & 93.3 & 68.4 & \cellcolor{uheat!32}79.6 & \cellcolor{uheat!21}52.1 & \cellcolor{uheat!28}69.8 & 65.1 & 63.7 & 48.4 \\
Qwen3.8-27B (Thinking) & \textbf{89.3} & \textbf{95.0} & \textbf{83.3} & \cellcolor{uheat!37}\textbf{91.5} & \cellcolor{uheat!33}\textbf{81.7} & \cellcolor{uheat!32}\underline{80.2} & \textbf{80.1} & \textbf{80.4} & \textbf{67.5} \\
\midrule
Gemini 3.8 Flash$^\dagger$ & 94.4 & 92.0 & 96.9 & \cellcolor{uheat!40}99.8 & \cellcolor{uheat!38}96.0 & \cellcolor{uheat!38}95.9 & 88.8 & 90.3 & 81.3 \\
Gemini 4 Argon$^\dagger$ & 94.1 & 96.8 & 91.3 & \cellcolor{uheat!39}98.2 & \cellcolor{uheat!37}91.5 & \cellcolor{uheat!35}88.1 & 87.8 & 90.1 & 83.2 \\
GPT-6.1 Sol$^\dagger$ & 97.3 & 97.7 & 96.9 & \cellcolor{uheat!40}99.5 & \cellcolor{uheat!38}96.2 & \cellcolor{uheat!38}96.1 & 94.6 & 95.7 & 90.5 \\
\bottomrule
\end{tabular*}

\caption{\textbf{Main results on \bench{}} (accuracy, \%). Per query: Ord./Insuf. cover orderable/insufficient inputs, and Overall is their query-weighted mean. $U_0$/$U_1$/$U_2$ (shaded by rate): correct-UNCLEAR rate on insufficient inputs with zero, one, or both target actions evidenced. Joint: both inputs of a pair (\PairHL{}, \PairKP{}) or all eight queries of a family (Family) correct together. Rows: fixed-output baselines, open instruction models and Thinking configurations, closed references ($^\dagger$). Bold/underline: best/second-best open configuration. 95\% intervals reach at most 4.5 points from the estimate for per-query and joint metrics and 9.4 for the type columns (\cref{app:intervals}).}
\label{tab:main}
\end{table*}
\section{Experiments}\label{sec:experiments}
We ask three questions: where open models fall short, when they answer insufficient inputs, and what narrows the gap. In short, the best open models order nearly as accurately as the closed references but withhold far less often (\cref{sec:exp_main}); they withhold mainly when neither action is evidenced, across and within families, and otherwise answer with informed guesses (\cref{sec:exp_analysis}); and Thinking narrows the gap without closing it (\cref{sec:exp_improve}).

\subsection{Experimental Setup}
We evaluate sixteen open-weight configurations: Qwen3-VL-8B/32B-Instruct and their Thinking variants~\cite{qwen3vl}, InternVL3.5-8B/14B~\cite{internvl35}, MiMo-VL-7B-SFT/RL~\cite{mimovl} and MiMo-VL-7B-RL with Thinking switched on, LLaVA-OneVision-2-8B~\cite{onevision2}, Gemma~4~12B~\cite{gemma4}, GLM-4.6V-Flash~\cite{glm46v}, Keye-VL-1.5-8B~\cite{keye15}, MiniCPM-V4.7~\cite{minicpmv47}, and Qwen3.8-27B~\cite{qwen38} with Thinking switched off and on. Pairs within a series differ in scale, post-training, or Thinking; models with a thinking switch otherwise run without Thinking. A configuration is a checkpoint with its thinking setting; we call the twelve that run without Thinking \emph{instruction models} and the four that run with it \emph{Thinking configurations}. As closed-source references we query Gemini 3.8 Flash, Gemini 4 Argon, and GPT-6.1 Sol, with reasoning enabled, through their APIs with the same images and prompts. All configurations also answer the queries of the matched diagnostic subset of \cref{sec:exp_analysis}.

Frames are passed as separate images in chronological order with each model's native preprocessing and chat template, together with the two target descriptions and the three answer options; observation-set names and labels are never included. The prompt states that neither action is guaranteed to occur, that an unseen action must not be assumed to happen earlier or later, and that UNCLEAR is required unless the images establish the order of both actions, so a model that answers on an insufficient input does so against an explicit instruction (\cref{fig:prompt}). Decoding is greedy, with at most 256 output tokens for instruction models and 4,096 for Thinking configurations. One rule, blind to the reference labels, extracts a single final label for every model; invalid and truncated outputs count as errors. Checkpoints, API settings, and the verbatim prompts are given in \cref{app:protocol}.

\subsection{Where Do Open Models Fall Short?}\label{sec:exp_main}
\begin{figure}[t]
\centering
\includegraphics[width=\linewidth]{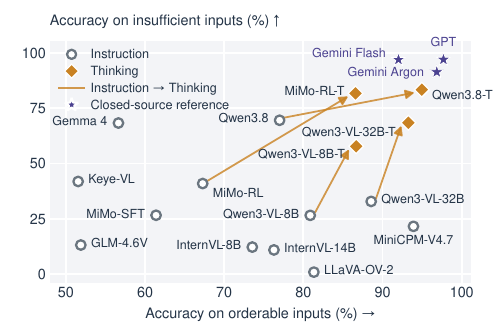}
\caption{\textbf{Ordering accuracy does not predict withholding.} Accuracy on orderable vs.\ insufficient inputs; arrows connect each instruction model to its Thinking configuration (-T).}
\label{fig:scatter}
\end{figure}
\begin{figure*}[t]
\centering
\includegraphics[width=\textwidth]{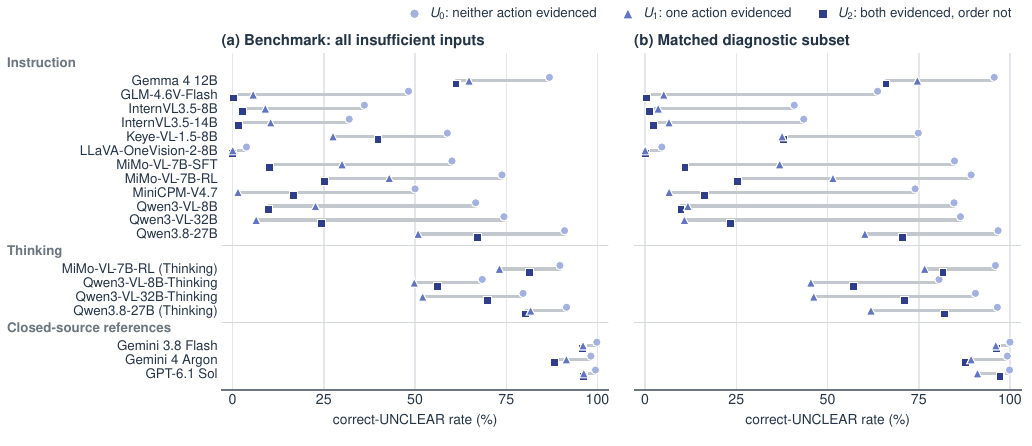}
\caption{\textbf{Withholding depends on what is missing, across and within families.} Correct-UNCLEAR rate by evidence type (a) on the benchmark and (b) on the matched diagnostic subset, whose inputs of a family share the episode and question. Each row spans the three types of one configuration; configurations are listed in the order of \cref{tab:main}. Differences quoted in the text are paired within families (\cref{app:matched}).}
\label{fig:types}
\end{figure*}

The best open models order nearly as accurately as the closed references but withhold far less often (\cref{tab:main}). On orderable inputs, the setting that forced-choice benchmarks evaluate, the best open configurations fall within the range of the references. On insufficient inputs, the strongest open configuration, Qwen3.8-27B with Thinking, withholds correctly on 83.3\% of queries, whereas every reference exceeds 90\%, and ten of the twelve instruction models stay below 50\%.

Ordering accuracy does not predict withholding (\cref{fig:scatter}). Across open configurations the two accuracies are only weakly correlated (Pearson $r=0.26$), and among the instruction models alone the correlation is negative ($r=-0.26$): MiniCPM-V4.7 is among the best at ordering yet among the weakest at withholding, whereas Gemma~4 withholds often on insufficient inputs but also on 31.1\% of orderable ones.

Overall accuracy hides this gap, and joint metrics expose it (\cref{tab:main}). Because always answering UNCLEAR already reaches 48.7\% overall, six open configurations score below this constant output. When all eight queries of a family must be correct, Qwen3.8-27B with Thinking succeeds on two thirds of the families and the best instruction model, the same model without Thinking, on a third, whereas the closed references succeed on 81--91\%.

Two checks show that this gap reflects the evidence in the frames. First, it is not an artifact of input length. Where answerability changes with the frame count~\cite{virtuebench}, insufficient inputs are also shorter, but $H$ and $L$ contain the same number of frames from the same episode, and MiniCPM-V4.7 still answers 92.3\% of $H$ queries but only 24.7\% of $L$ queries correctly. Second, the distinction can be drawn from the frames alone. The withholding of the closed references follows the label rather than the observation set: they answer UNCLEAR on 1--9\% of the 90 queries whose $L$ or $P$ input is orderable, against 92--98\% of those whose $L$ or $P$ input is insufficient.

\subsection{When Do Models Answer Insufficient Inputs?}\label{sec:exp_analysis}

\begin{figure*}[t]
\centering
\includegraphics[width=\textwidth]{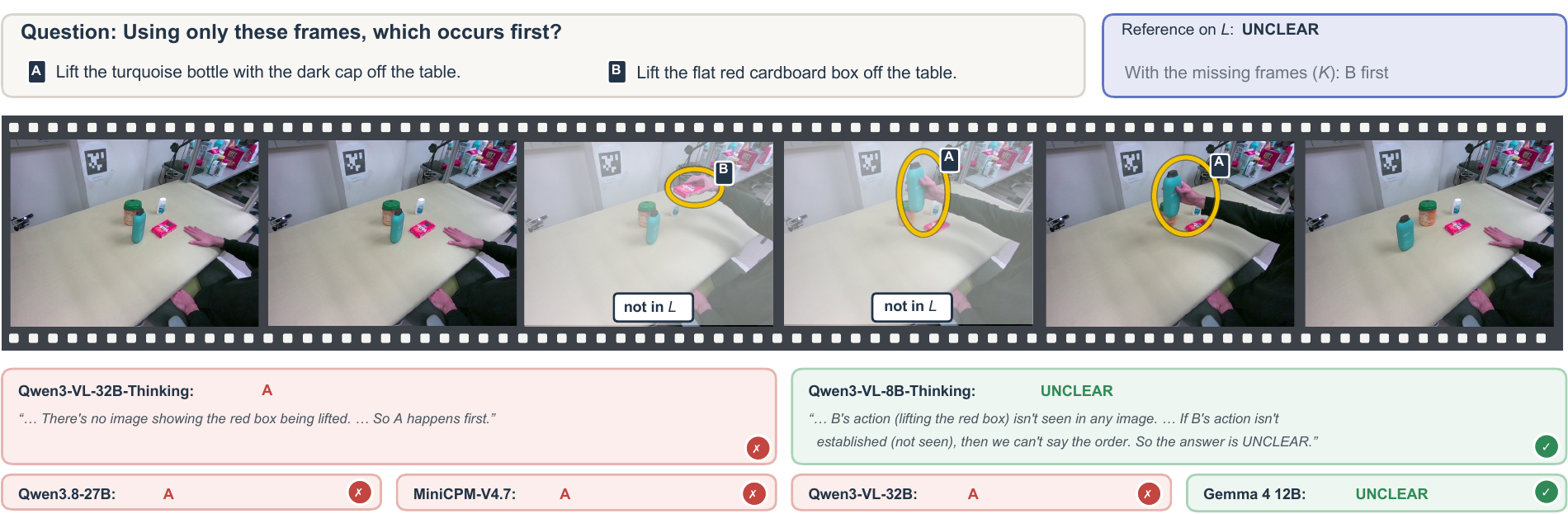}
\caption{\textbf{Ordering an action that no frame shows.} On this $U_1$ input, $L$ evidences A but not B, yet four of the six models shown answer that A happened first. The dimmed frames, added in $K$, show B before A. Quotes are verbatim; random examples are in \cref{app:case}.}
\label{fig:case}
\end{figure*}

\begin{figure}[!t]
\centering
\includegraphics[width=\columnwidth]{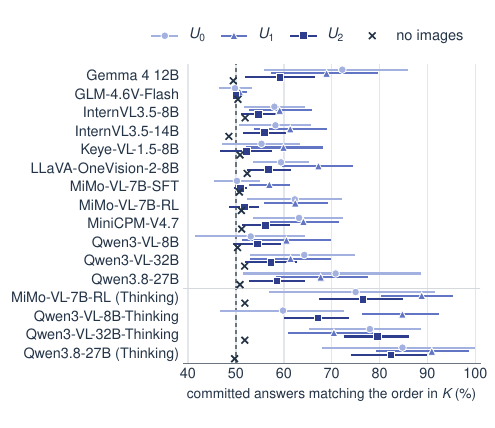}
\caption{\textbf{Committed answers on insufficient inputs that match the order shown in $K$} (\%, 95\% cluster-bootstrap intervals). The dashed line marks 50\%, the agreement of a fixed letter preference over complete pairs; crosses give the agreement when the model sees only the target descriptions, without images (not run for Qwen3-VL-8B-Thinking).}
\label{fig:committed}
\end{figure}

Most open models thus answer more insufficient inputs than they withhold. Do they answer all of them alike, or does it depend on what is missing? It depends: they answer mainly once the frames evidence at least one action (\cref{fig:types}). Every open configuration withholds most reliably when neither action is evidenced ($U_0$) and less often once one action ($U_1$) or both ($U_2$) are evidenced; the closed references follow the same pattern at a much higher level. Which kind of partial evidence draws more answers depends on the model: six configurations withhold significantly less often on $U_1$ than on $U_2$, and six others the reverse.

The same pattern holds when the episode and question are fixed. Because evidence types mostly come from different families (\cref{sec:stats}), we build a \emph{matched diagnostic subset}: 1,053 further four-frame inputs for 400 of the families, which keep the episode and question and change only which actions the frames evidence. The new inputs are built to evidence neither action or exactly one, are labeled under the same criteria but reviewed by one annotator, and do not enter the metrics of \cref{tab:main} (\cref{fig:matched}; \cref{app:matched}). On this subset (\cref{fig:types}b), evidence of one action lowers correct withholding relative to evidence of neither for every configuration: by 19--75 points for the open models except LLaVA-OneVision-2-8B, which almost never withholds, and by only 4--9 points for the closed references. One evidenced action is often enough to turn a correct withholding into an answer: where they answered UNCLEAR on a family's $U_0$ inputs, seven of the twelve instruction models order 87--100\% of its $U_1$ inputs. Open models thus appear to answer when the frames are \emph{relevant} to the question, not when they are \emph{sufficient} for it.

Models commit even when they correctly report what is missing, instruction models most of all. In \cref{fig:case}, Qwen3-VL-32B-Thinking states that no frame shows B and then answers that A came first, although the prompt forbids ordering an unseen action; reasoning language models likewise voice uncertainty and still answer~\cite{abstentionbench}. Occurrence queries show the same at scale. On $U_1$ inputs where an instruction model itself judges correctly that only one action is established, five of six still order the two actions in 69--100\% of cases; on $U_2$ inputs where a model judges both actions established, the five with enough such inputs fail to withhold in 62--100\% (\cref{app:occurrence}). The orders they give favor the action they can see: the evidenced action comes first in only 52.6\% of $U_1$ inputs, yet committed answers put it first in 52--88\% of cases, significantly above 50\% for eleven of the sixteen configurations and below it for none (\cref{app:committed}).

Committed answers are nevertheless informed guesses. Because labels are swapped, a fixed letter preference agrees with the order shown by the family's $K$ input only about half of the time, yet committed answers agree significantly more often in 34 of the 48 combinations of configuration and evidence type (\cref{fig:committed}). In a control without images, where models must choose from the target descriptions alone, the fifteen open configurations we ran agree with $K$ on only 48.6--52.4\% of queries (\cref{app:committed}). For the thirteen configurations whose estimates can be interpreted, keeping only content-based choices, in which a model picks the same action under both letter assignments, the target descriptions alone put the true first action first in 41.2--63.0\% of families, every interval including 50\%, whereas the committed content-based answers on insufficient inputs do so in 58.0--83.9\%, every interval above 50\%. The frames thus contain cues that favor the true order without establishing it, and a forced-choice benchmark would credit these guesses as temporal understanding.

Added frames change answers only when they carry the missing evidence. Models that correctly withheld on $L$ order the family's $K$ input about as accurately as $K$ inputs in general (at most five points lower for fifteen of the sixteen configurations), whereas the two peripheral frames of $P$ leave accuracy within five points of that on $L$ for fifteen. The bottleneck is thus withholding on $L$: where a model fails on $L$ or on $K$ of the same family, the error lies mostly on $L$, for every configuration except Gemma~4 (\cref{app:added}).

\subsection{What Narrows the Gap?}\label{sec:exp_improve}
\begin{table}[t]
\centering\footnotesize\setlength{\tabcolsep}{3.2pt}
\begin{tabular*}{\linewidth}{@{\extracolsep{\fill}}lrrrrr@{}}
\toprule
Change & Ord. & $U_0$ & $U_1$ & $U_2$ & Family \\
\midrule
Qwen3-VL 8B$\to$32B & +7.7 & +7.8 & $-$16.2 & +14.5 & +6.4 \\
InternVL3.5 8B$\to$14B & +2.7 & \textcolor{gray}{$-$4.2} & \textcolor{gray}{+1.5} & \textcolor{gray}{$-$1.1} & \textcolor{gray}{+0.0} \\
\midrule
MiMo-VL SFT$\to$RL & +5.9 & +13.7 & +12.9 & +15.2 & +4.8 \\
\midrule
MiMo-VL-RL$\to$Thinking & +19.3 & +15.9 & +30.2 & +56.1 & +40.8 \\
Qwen3-VL-8B$\to$Thinking & +5.8 & \textcolor{gray}{+1.8} & +27.1 & +46.1 & +15.4 \\
Qwen3-VL-32B$\to$Thinking & +4.7 & +5.2 & +45.6 & +45.3 & +25.2 \\
Qwen3.8-27B$\to$Thinking & +18.0 & \textcolor{gray}{+0.5} & +30.8 & +13.1 & +34.5 \\
\bottomrule
\end{tabular*}

\caption{\textbf{Paired changes within model series} (points, same queries). Thinking raises withholding most where one or both actions are evidenced. MiMo-VL-RL and Qwen3.8-27B switch Thinking on in the same checkpoint; the other pairs are separately released checkpoints. Gray: 95\% cluster-bootstrap interval includes zero; full intervals in \cref{app:checkpoints}.}
\label{tab:comparisons}
\end{table}

If partial evidence is what draws an answer, what makes a model resist it? Thinking does, above all where partial evidence invites an answer (\cref{tab:comparisons}). In every Thinking pair, Thinking raises correct withholding more on $U_1$ and $U_2$ than on $U_0$, and it also raises orderable accuracy; for Qwen3-VL-32B, for example, it adds 46 and 45 points on $U_1$ and $U_2$ against 5 on $U_0$. It does not make models uniformly more cautious: UNCLEAR answers on orderable inputs rise by at most 4 points and fall for Qwen3.8-27B.

Thinking thus separates sufficient from insufficient evidence more sharply, and what a model reports then carries over to its answer: on $U_1$ inputs where they identify the missing action in a separate occurrence query, Qwen3-VL-32B and Qwen3.8-27B order the actions in 98\% and 69\% of cases without Thinking but in only 42\% and 13\% with it.

Because two pairs, MiMo-VL-7B-RL and Qwen3.8-27B, differ only in the official thinking switch, this effect is not specific to separately trained Thinking checkpoints; the other pairs compare released checkpoints that also differ in training data and recipe. Earlier findings are mixed: reasoning fine-tuning lowered abstention in language models~\cite{abstentionbench}, and on long videos reasoning variants mostly refused more accurately, but not Qwen3-VL-32B~\cite{virtuebench}; our gains may arise because the prompt states exactly what an answer requires, which reasoning can check.

Thinking does not remove the pull of partial evidence. On the matched subset, its remaining failures concentrate in $U_1$ inputs that begin after the first action, so that their first frame already shows the result of that action. The question asks what occurs within the segment, and a result that is already present does not establish that its action occurred there (\cref{sec:task_definition}); the closed references accordingly withhold on 80--94\% of these queries. The Thinking configurations withhold on only 11--57\%, against 61--86\% on other $U_1$ inputs, and their reasoning infers an order from the visible result: ``If red had been placed first, red would be present in frame~1. It's not. So green must have been placed before red'' (Qwen3.8-27B with Thinking; \cref{app:matched}). With these inputs, Thinking does not raise matched $U_1$ withholding for Qwen3.8-27B; without them, it does so in all four pairs. Reasoning thus cuts both ways: it helps a model check what the frames show, but it also lets the model infer an order from a visible result and answer with it, although the question concerns only what the frames establish.

Scale and post-training act differently. Scaling Qwen3-VL from 8B to 32B raises accuracy on insufficient inputs by 6.3 points (\cref{tab:main}), but this total combines opposite changes (\cref{tab:comparisons}): withholding rises on $U_2$ by about 15 points and falls on $U_1$ by about 16, so the larger model commits more often once one action is evidenced; scaling InternVL3.5 from 8B to 14B changes only orderable accuracy. RL post-training of MiMo-VL, by contrast, raises withholding on every type. Changing the prompt does not close the gap either: for the four models we re-prompted, no prompt variant improves both accuracies, and the most explicit one lifts Gemma~4 to 94.5\% on insufficient inputs only by cutting it to 18.8\% on orderable ones (\cref{app:strategies}).

\section{Limitations}\label{sec:limitations}
\paragraph{Scope.} \bench{} targets the order of two actions in a few frames, which are selected to control the evidence; how often a sampling strategy yields such inputs is not measured. It does not evaluate long-video understanding, causal reasoning, or how a system should choose additional frames.

\paragraph{Comparisons between evidence types.} On the benchmark, evidence types mostly differ between families, and the type is partly tied to the kind of action, as handle actions rarely leave a lasting trace. The matched subset compares types within families, but its inputs still differ in which frames, and which part of the episode, they show, and its families come mostly from Perception Test.

\paragraph{Inputs and models.} Equal frame counts remove input length as an explanation, yet $H$ and $L$ can still differ in how easy their frames are to perceive, and models encode the same frames with different numbers of visual tokens. The overlap between the benchmark videos and the models' training data is unknown. The closed-source references were sampled once per query, and their versions may change.

\section{Conclusion}\label{sec:conclusion}
We asked when sampled frames establish the order of two actions, and built \bench{} to test whether models answer only then: event families that vary only the frames, labels that record what each input establishes, and metrics that require ordering and withholding together. The best open-weight models order nearly as accurately as the closed-source references, yet they answer when the frames are relevant rather than sufficient: one visible action is usually enough. Thinking narrows this gap, but reasoning cuts both ways, since it also lets a model infer an order from a visible result that the frames do not establish. Temporal understanding should therefore be scored by whether an answer is supported by the frames a model was given, not only by whether it agrees with the full video. Because every insufficient input records what is missing, \bench{} can also score the evidence a model reports as missing and guide methods that request further frames instead of guessing. Next steps are labels for individual frames, questions about more than two actions, and robot and world-model settings, where an agent must know what its sparse observations establish before acting.

\section*{LLM Usage Statement}
We used OpenAI Codex and Anthropic Claude as AI assistants to write and organize code for data processing, evaluation, analysis, and figures; to search the literature; and to draft and edit the manuscript. Claude alone was used to pre-screen candidate episodes, to draft candidate questions, frame sets, and provisional labels, and to record descriptive analysis metadata such as object visibility, which we spot-checked. At least two human annotators reviewed every benchmark item and corrected labels and frames where needed; the labels of the matched diagnostic subset were also drafted with Claude, and one annotator reviewed all of them. Model outputs were scored by a fixed rule, not by an AI judge. The authors take full responsibility for the content of the paper, including its citations and numerical claims.

{\small\bibliographystyle{ieeenat_fullname}\bibliography{references}}

\clearpage
\appendix
\counterwithin{figure}{section}\counterwithin{table}{section}   
\appendixtables
\renewcommand{\floatpagefraction}{0.8}\renewcommand{\dblfloatpagefraction}{0.8}
\section*{Appendix}
The appendix describes how \bench{} was built (\cref{app:data}) and evaluated (\cref{app:protocol}), gives additional results in the order of \cref{sec:experiments} (\cref{app:robustness}), shows randomly drawn examples (\cref{app:case}), and documents provenance and release (\cref{app:release}).

\section{Benchmark Construction}\label{app:data}
\subsection{Construction Pipeline}\label{app:review}
\begin{figure*}[!t]
\centering
\includegraphics[width=\textwidth]{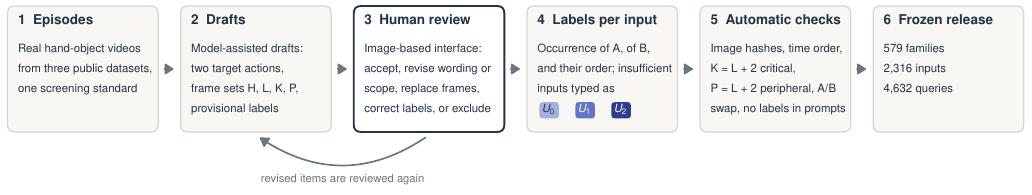}
\caption{\textbf{Construction of \bench{}.} Drafting is model-assisted; every item is reviewed by at least two human annotators, and revised items are reviewed again before automatic checks and release.}
\label{fig:pipeline}
\end{figure*}

\Cref{fig:pipeline} summarizes how a family is built. Candidate episodes come from three public datasets of real hand--object interaction and are screened under one standard. For each episode, Claude, which is not among the evaluated models, drafts two target actions, the four frame sets, and provisional labels from candidate frames. At least two human annotators then review every item in an image-based interface that shows the ordered frames, the target descriptions, and the labels. Annotators can accept an item, revise its wording or scope, replace frames so that each input serves its intended role, correct its labels, or exclude it; revised items return for another review, and accepted corrections replace the earlier annotation. An item is released only when all annotators accept its wording, frames, and labels. Before release, annotators excluded 18 of the 597 reviewed families (3.0\%), all because poor video quality, such as blur or low light, made it impossible to tell, even from the full source video, whether an action occurred; frames that leave the order open lead to an UNCLEAR label, not to exclusion (\cref{app:annotation}). On 32 randomly drawn families, two annotators labeling independently and without the provisional labels agreed on all occurrence and order labels for 125 of the 128 inputs (97.7\%). Reviewed scope statements enter the prompts (\cref{fig:prompts_main}c); review notes never do.

Automatic checks complement the review: they verify image hashes, chronological presentation, the set relations $K=L+2$ critical and $P=L+2$ peripheral frames, label consistency under the A/B swap, and that no label, note, or set name enters a model request.

\subsection{Annotation Criteria}\label{app:annotation}
\paragraph{Event specification.}
Each target names an object, an action, and a recipient where applicable. The two descriptions can be read independently, and any spatial reference or temporal scope that is needed is stated in the question. The event boundary follows the reviewed description: it may be a specified result, a visible operation, or a contact, rather than the start or end of a whole activity. When an action may repeat, a scope statement selects its first specified occurrence.

\paragraph{Occurrence evidence.}
An occurrence is established by a visible operation or by a state change that can only arise from the specified event. A completed state alone does not show that it formed within the segment, whereas an informative transition can establish occurrence without the action itself being seen. Identity and state may be judged from all frames of the current input, so partial occlusion does not invalidate an item; evidence from another input or an unsampled interval is never used. NOT\_ESTABLISHED means that the input does not support the occurrence, not that the event did not happen. As a controlled benchmark, \bench{} labels only what the given frames establish and never guesses beyond them. The judgments rest on three explicit assumptions: the frames are in temporal order; a state persists between consecutive frames unless a frame shows otherwise, so no unobserved reversal or repetition between two frames is assumed; and where an action could repeat, the scope statement fixes which occurrence is meant. What counts as evidence follows the event boundary of the description: a specified result can be established by the state change it produces, whereas a specified operation or contact must itself be visible. Sparse frames cannot exclude a repetition hidden inside an unsampled interval, so a label states what the frames establish under these assumptions, not a complete record of the episode.

\paragraph{Order evidence.}
A definite order requires both occurrences and their strict relation. When both initial states precede both final states but no frame resolves the intermediate order, the reference answer stays UNCLEAR (\cref{fig:overview}). A definite order needs a frame in which one action is complete and the other has not yet occurred; the arrangement of the final states, such as shoes resting on a folded garment, does not by itself show which change came first. The first frame that shows two completed outcomes does not establish simultaneity. Conversely, overlapping activities can have ordered event boundaries if the specified transitions are identifiable.

\paragraph{Item validity.}
Missing occurrence or order evidence is compatible with a valid question. Ambiguity in the target, recipient, or event definition is not: it changes the proposition being evaluated and leads to revision or exclusion. This separates uncertainty in the evidence from a poorly posed question. Labels are assigned from the reviewed input, independently of its intended role.

\subsection{Labels Across Observation Sets}\label{app:family_annotation}
\begin{figure*}[!t]
\centering
\includegraphics[width=\textwidth]{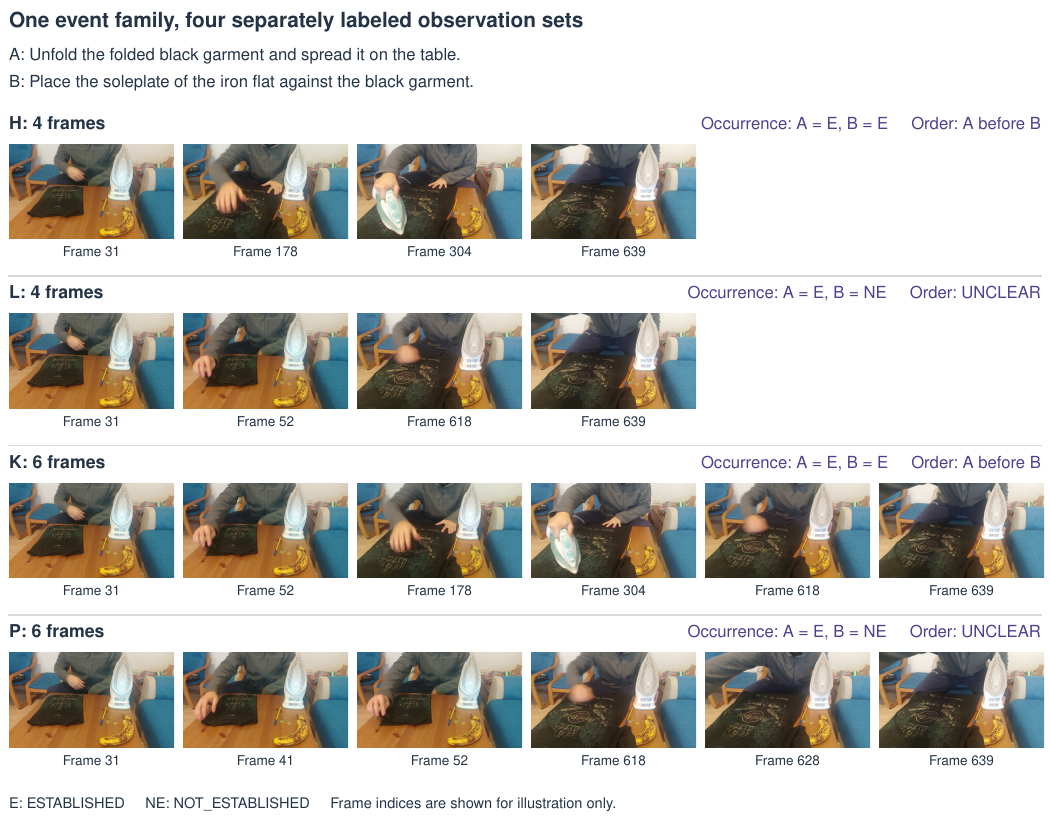}
\caption{\textbf{Labels of all four observation sets of one family.} Frames are shown in temporal order at the same scale. Each row gives the separately reviewed occurrence labels for A and B and the order label. NOT\_ESTABLISHED denotes insufficient evidence, not confirmed absence. Frame indices and labels are annotations for the reader and are not part of the model inputs.}
\label{fig:annotation_family}
\end{figure*}

Every set is labeled on its own: occurrence of A, occurrence of B, and their order are judged from the images of that set alone, and a six-frame set does not inherit the label of the four-frame set it extends. Review is organized accordingly: each review batch contains a single observation set, such as only $H$ inputs or only $L$ inputs, so no item is shown with the other sets of its family. \Cref{fig:annotation_family} shows a complete family. Unfolding is established in all four sets, whereas soleplate contact is established only in $H$ and $K$, so the order is A before B in $H$ and $K$ and UNCLEAR in $L$ and $P$. The two critical frames in $K$ supply the missing contact evidence; the two peripheral frames in $P$ do not. This is one reviewed example rather than a prescribed pattern: in 30 control families the labels depart from it, mostly because all four sets support the same answer (\cref{app:units}).

\subsection{Event Scope}\label{app:boundary_examples}
\begin{figure*}[!t]
\centering
\includegraphics[width=\textwidth]{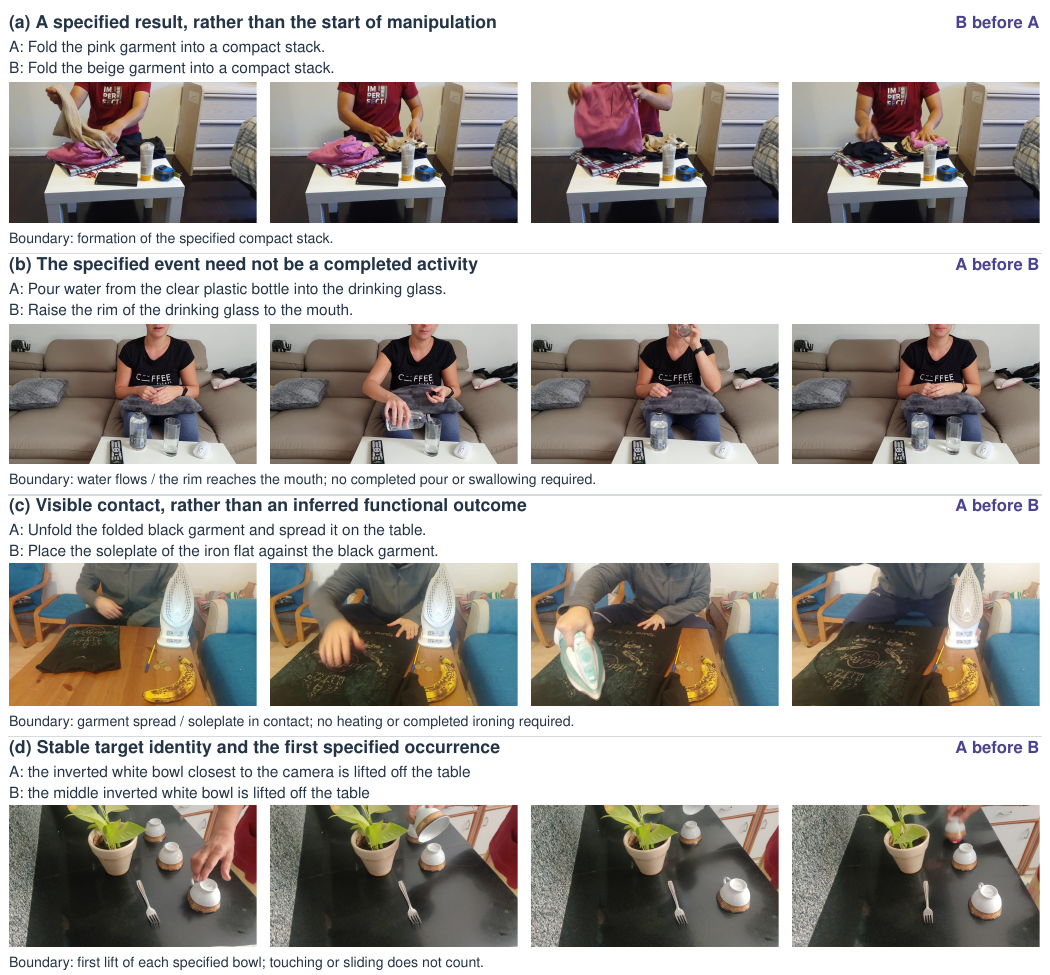}
\caption{\textbf{Event boundaries in the reviewed annotations.} Each row is a complete four-frame $H$ input in temporal order with its targets, the relevant boundary, and the reference order: (a) a specified result; (b) a specified operation; (c) visible contact without an inferred outcome; (d) target identity and first occurrence. Target descriptions are verbatim; the boundary lines summarize the scope statements, which appear verbatim in the prompts. Images are unaltered apart from resizing.}
\label{fig:annotation_boundaries}
\end{figure*}

\Cref{fig:annotation_boundaries} shows four event definitions. In (a), the outcome is a compact folded stack; first contact or ongoing folding does not count. In (b), water entering the glass and the rim reaching the mouth define the events, without requiring the pour to finish or the person to swallow. In (c), unfolding is compared with visible soleplate contact rather than with inferred heating. In (d), the targets are identified by their initial positions and tracked to their first lifts; touching the bowls is not enough. A single onset or completion rule for all events would change these questions, which is why scope is stated per family.

\subsection{Frame Timing and Composition}\label{app:stats}

\begin{figure*}[t]
\centering
\includegraphics[width=\textwidth]{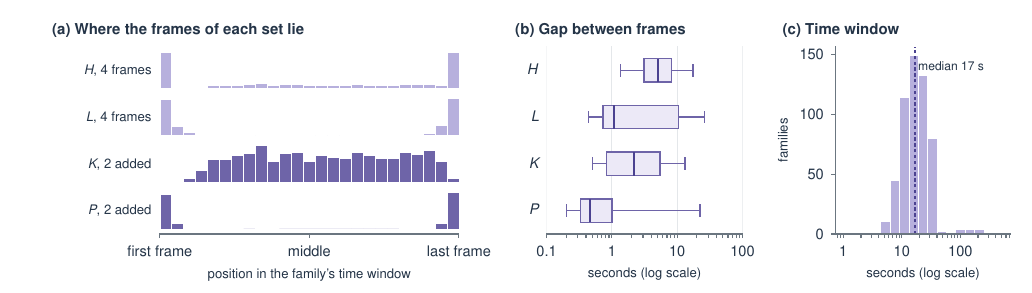}
\caption{\textbf{Temporal structure of the observation sets} (549 families with frame timestamps). (a) Position of each set's frames within the family's time window, from its first to its last sampled frame; for $K$ and $P$ only the two frames they add to $L$ are shown. (b) Gap between consecutive frames of one input (boxes: quartiles; whiskers: 5th--95th percentiles). (c) Length of the time window.}
\label{fig:timing}
\end{figure*}
\begin{figure*}[!t]
\centering
\includegraphics[width=\textwidth]{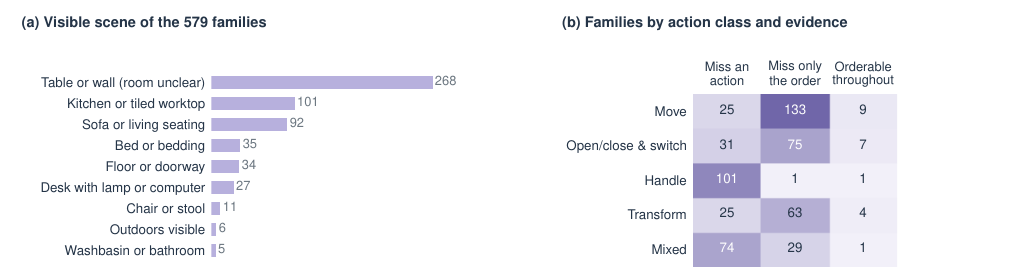}
\caption{\textbf{Composition of \bench{}.} (a) Visible scene of each family (analysis metadata). (b) Families by action class (the shared class of the two targets, otherwise Mixed) and by the evidence of their insufficient inputs; 22 families are orderable in all four sets.}
\label{fig:composition}
\end{figure*}
\Cref{fig:timing} shows how the four sets sample time, for the 549 families whose source videos provide frame timestamps. A family spans a median of 17\,s from its first to its last sampled frame. $H$ spreads its four frames over this window, with consecutive frames a median of 5.1\,s apart. $L$ places two frames near each end of the window (545 of the 549 families) and leaves the middle unsampled, so both actions can complete between its second and third frames. The critical frames added in $K$ fall inside this gap, where the actions happen, whereas the peripheral frames added in $P$ sit next to frames already in $L$; consecutive frames of $P$ are a median of 0.5\,s apart. Equal frame counts therefore come with very different coverage of the interval in which the order is decided.

\Cref{fig:composition} shows the visible scenes and how action classes relate to evidence. Scene categories describe the visible context and are analysis metadata, not labels. Handle actions, which rarely leave a lasting trace, are the largest class in families whose insufficient inputs miss an action (247 of their 512 targets), and 101 of the 103 families whose two targets are handle actions are of this kind.

\subsection{Corpus and Scoring Units}\label{app:units}
The 579 families, 532 from Perception Test, 30 from HO-Cap, and 17 from CaptainCook4D, contain 2,316 inputs and 4,632 order queries; the original and swapped assignments use the same ordered images, and every query is independent rather than a turn in a conversation. In 549 standard families $H$ and $K$ are orderable while $L$ and $P$ are insufficient. The other 30 keep whatever their frames support: 22 are orderable throughout, seven insufficient throughout, and one has orderable $H$, $K$, $P$ but insufficient $L$. Insufficient inputs comprise 277 in $U_0$, 240 in $U_1$, and 610 in $U_2$, and no family has both inputs that miss an action ($U_0$, $U_1$) and inputs that miss only the order ($U_2$). In nine families $L$ is in $U_0$ and $P$ in $U_1$; in two control families the type also differs between sets, with $H$ and $K$ in $U_1$ and $L$ and $P$ in $U_0$. Because the observation set predicts the label in the standard families, the orderable $L$ and $P$ inputs of the control families show whether withholding follows the label or the set. On their 90 queries, Gemini 3.8 Flash, Gemini 4 Argon, and GPT-6.1 Sol answer UNCLEAR on 8.9\%, 1.1\%, and 6.7\%, far below their rates on insufficient $L$ and $P$ queries (97.5\%, 91.8\%, and 97.3\%). Family uses all families, \PairHL{} and \PairKP{} use 550 and 549, the critical addition uses 550, and the label-preserving peripheral addition uses 578.

All results come from one evaluation of the full collection. Intervals use a cluster bootstrap with 5,000 resamples over the 558 known recording and participant groups (seed 20261106), keeping all queries of a group together; ratios stay query- or family-weighted as defined by each metric. Intervals describe this collection, and paired comparisons are not corrected for multiplicity.

\section{Evaluation Protocol}\label{app:protocol}
\begin{table*}[!t]
\centering\scriptsize\setlength{\tabcolsep}{3.5pt}
\begin{tabular}{lllrr}
\toprule
Model & Checkpoint & Revision & Tokens & Hours \\
\midrule
Gemma 4 12B & \nolinkurl{google/gemma-4-12B-it} & {\tiny\texttt{707f0a3b8a3c7ad586ed01e27eafbad8a27dd0f7}} & 13,286 & 0.27 \\
GLM-4.6V-Flash & \nolinkurl{zai-org/GLM-4.6V-Flash} & {\tiny\texttt{411bb4d77144a3f03accbf4b780f5acb8b7cde4e}} & 18,828 & 0.68 \\
InternVL3.5-8B & \nolinkurl{OpenGVLab/InternVL3_5-8B} & {\tiny\texttt{9bb6a56ad9cc69db95e2d4eeb15a52bbcac4ef79}} & 9,555 & 0.92 \\
InternVL3.5-14B & \nolinkurl{OpenGVLab/InternVL3_5-14B} & {\tiny\texttt{a1e37197b393ce9eec9df700fef65c11f4a6ffbd}} & 9,527 & 1.30 \\
Keye-VL-1.5-8B & \nolinkurl{Kwai-Keye/Keye-VL-1_5-8B} & {\tiny\texttt{b4d413e4f42a3999af1f5ba0e281c7d51d6b93de}} & 10,806 & 0.89 \\
LLaVA-OneVision-2-8B & \nolinkurl{lmms-lab-encoder/LLaVA-OneVision-2-8B-Instruct} & {\tiny\texttt{878cde7e04fd962a6082951d8f83083a73fea588}} & 9,285 & 0.46 \\
MiMo-VL-7B-SFT & \nolinkurl{XiaomiMiMo/MiMo-VL-7B-SFT-2508} & {\tiny\texttt{11e8fdc278f5e1e6a72f9561772a6d7cd6921d33}} & 37,743 & 0.95 \\
MiMo-VL-7B-RL & \nolinkurl{XiaomiMiMo/MiMo-VL-7B-RL-2508} & {\tiny\texttt{ec4ebdc29b83b94aa7f62670c78376250dad8dbc}} & 38,194 & 0.95 \\
MiniCPM-V4.7 & \nolinkurl{openbmb/MiniCPM-V-4.7-35B-A3B} & {\tiny\texttt{98e79ba972b586ac6146d3505771ce1467e199ba}} & 9,756 & 0.76 \\
Qwen3-VL-8B & \nolinkurl{Qwen/Qwen3-VL-8B-Instruct} & {\tiny\texttt{0c351dd01ed87e9c1b53cbc748cba10e6187ff3b}} & 9,898 & 0.40 \\
Qwen3-VL-32B & \nolinkurl{Qwen/Qwen3-VL-32B-Instruct} & {\tiny\texttt{0cfaf48183f594c314753d30a4c4974bc75f3ccb}} & 10,033 & 1.05 \\
Qwen3.8-27B & \nolinkurl{Qwen/Qwen3.8-27B} & {\tiny\texttt{1d4bf0f2ff6012fd82039f2fa52739d0dd7c60c0}} & 11,119 & 1.51 \\
MiMo-VL-7B-RL (Thinking) & \nolinkurl{XiaomiMiMo/MiMo-VL-7B-RL-2508} & {\tiny\texttt{ec4ebdc29b83b94aa7f62670c78376250dad8dbc}} & 4,623,472 & 23.80 \\
Qwen3-VL-8B-Thinking & \nolinkurl{Qwen/Qwen3-VL-8B-Thinking} & {\tiny\texttt{92f3c4b4feadd3a016ef468d103bb5f58b2a2c6b}} & 4,249,403 & 26.17 \\
Qwen3-VL-32B-Thinking & \nolinkurl{Qwen/Qwen3-VL-32B-Thinking} & {\tiny\texttt{7edd10ffd1196091948fb245ff63e406ccb2d4d1}} & 2,248,581 & 25.01 \\
Qwen3.8-27B (Thinking) & \nolinkurl{Qwen/Qwen3.8-27B} & {\tiny\texttt{1d4bf0f2ff6012fd82039f2fa52739d0dd7c60c0}} & 2,511,338 & 45.52 \\
\bottomrule
\end{tabular}

\caption{\textbf{Evaluated checkpoints and compute.} Full pinned revisions; Qwen3-VL models without the Thinking suffix are the Instruct checkpoints, and MiniCPM-V4.7 is the 35B-A3B checkpoint. Tokens and Hours: generated tokens and summed generation time over the 4,632 first-pass queries.}
\label{tab:models}
\end{table*}

\subsection{Models and Inference}\label{app:models}
\Cref{tab:models} lists the checkpoints. Inference uses BF16, greedy decoding, seed 0, and SDPA attention on H100 GPUs with Torch 2.13.0 and Transformers 5.18.0; the two InternVL3.5 models use the Transformers versions their native code requires (4.52.1 and 4.57.6), and Keye-VL-1.5-8B, whose native code requires FlashAttention, runs with Torch 2.8.0, Transformers 4.55.4, and FlashAttention 2.8.3 while attention stays SDPA. \Cref{tab:models} also lists the generated tokens and summed generation hours of each first pass (output limit 256 tokens, 4,096 for Thinking configurations); time excludes loading and queueing and is not a controlled speed comparison. Each model uses its pinned native processor and chat template. Frames are passed as separate images rather than as a video: video processors may resample or merge frames, whereas image inputs keep exactly the labeled evidence. InternVL3.5 tiles each image at 448 pixels with up to 12 tiles plus a thumbnail, and MiniCPM-V4.7 uses at most nine slices with 16$\times$ downsampling. Models with a thinking switch run with it off: Gemma~4, GLM-4.6V-Flash, MiniCPM-V4.7, and Qwen3.8-27B through their chat templates' non-thinking setting, and both MiMo-VL checkpoints and Keye-VL-1.5-8B through their official \texttt{/no\_think} suffix. MiMo-VL-7B-RL with Thinking uses the same checkpoint without the suffix and a 4,096-token limit, and Qwen3.8-27B with Thinking uses the same checkpoint with the template's thinking setting at medium reasoning effort (the default effort adds a system message) and the same limit; without Thinking, its outputs never exceed 9 tokens, so the smaller limit never binds. Because preprocessing stays native, identical frames do not imply identical numbers of visual tokens. For models added after the initial set, 24 queries from three complete families checked execution and truncation before the full run; admission never depended on accuracy, and no query was filtered by it.

\subsection{Closed-Source References}\label{app:closed}
Gemini 3.8 Flash and Gemini 4 Argon were queried through the Gemini API on 9 November 2026 with the same 4,632 queries, images, prompts, and answer extraction as the open models, one sample per query and no system prompt. Thinking is on, with at most 4,096 output tokens including the reasoning. The thinking level is medium for Flash and high for Argon: at medium, Argon reasons for only 17 tokens per query on average (24-query check), against about 460 at high, whereas Flash reasons for about 510 at medium; both are comparable to Qwen3-VL-32B-Thinking. Images are sent at the API's default high resolution, and sampling parameters stay at their defaults. As for the open models, 24 queries checked execution before each full run. Flash answers every query; Argon reaches the output limit on 35 queries, which count as errors. GPT-6.1 Sol was queried through the OpenAI Responses API on the same date with the same queries, images, prompts, and answer extraction, one sample per query and no system or developer instruction. Reasoning effort is medium (the default), with at most 4,096 output tokens including the reasoning; images are sent at high detail, which keeps every frame at its original resolution. After the 24 check queries, the remaining queries ran with Flex processing, which uses the same model at a lower price; every response reports this tier. GPT-6.1 Sol answers every query within the limit.

\subsection{Queries and Prompts}\label{app:prompts}
\begin{figure*}[!t]
\centering
\includegraphics[width=\textwidth]{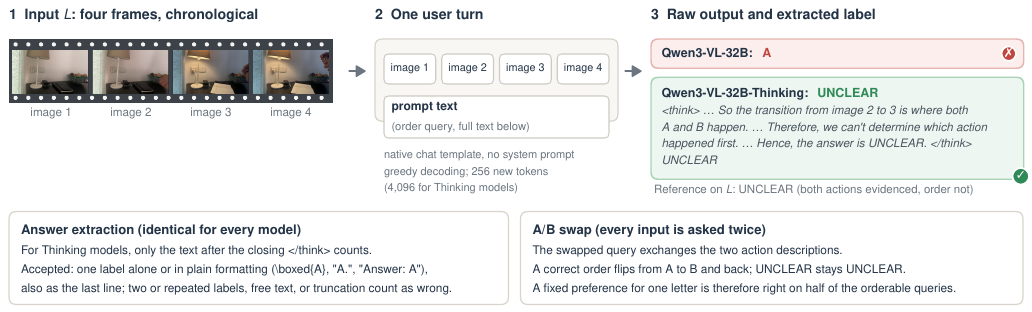}
\caption{\textbf{From frames to score.} Input $L$ of the family in \cref{fig:overview} (original assignment) with the real outputs of Qwen3-VL-32B and its Thinking variant. The reasoning is quoted verbatim, with omissions marked by ellipses.}
\label{fig:request}
\end{figure*}

\begin{figure*}[!t]
\centering\footnotesize
\begin{minipage}[t]{0.495\textwidth}\vspace{0pt}
\begin{promptbox}{(a) Order query: main prompt for all models}
Use only the supplied images as evidence. When there are multiple images, their presentation order is chronological, but they are sampled frames, not a complete recording.\par The question or statement does NOT guarantee that either action occurred. Do not assume that an unseen action happened earlier or will happen later. Seeing an object, a hand near it, or one action alone does not by itself establish the order of two actions. A static state can be used only when it actually establishes the relevant events and their order. Do not infer order from typical routines or object position alone.\par\smallskip Allowed answer labels: A, B, UNCLEAR. Choose A or B only when the supplied images establish the order of both specif{}ied actions. Otherwise choose UNCLEAR.\par Question: Using only the supplied images, which occurs f{}irst in the recorded segment: \slot{Turn on the table lamp with the beige shade.} (A), or \slot{Open the black-covered notebook.} (B)?\par\smallskip Return exactly one of the allowed answer labels above. Do not output an explanation, JSON, Markdown, or any other text.
\promptnote{Input $L$ of the family in \cref{fig:overview}, original assignment. Colored text is substituted per family; the rest is identical for every query, apart from the scope sentence in (c). MiMo-VL additionally ends the prompt with \texttt{/no\_think}.}
\end{promptbox}
\vspace{2pt}
\begin{promptbox}{(b) A/B swap: the only line that changes}
Question: Using only the supplied images, which occurs f{}irst in the recorded segment: \slot{Open the black-covered notebook.} (A), or \slot{Turn on the table lamp with the beige shade.} (B)?
\promptnote{The swapped query exchanges the two descriptions; a correct order flips (A$\leftrightarrow$B) and UNCLEAR stays UNCLEAR.}
\end{promptbox}
\end{minipage}\hfill
\begin{minipage}[t]{0.495\textwidth}\vspace{0pt}
\begin{promptbox}{(c) Question-specific scope (16 families)}
Question-specif{}ic scope: \slot{Use only the supplied sampled images, presented chronologically. Neither event is guaranteed to occur. Lack of evidence does not mean an event did not occur. Changes within the same sampling interval do not establish simultaneity. A is the f{}low of water from the specif{}ied plastic bottle into the glass; the entire pour need not be f{}inished. B is the glass rim reaching the mouth; swallowing or drinking is not required. Compare the f{}irst occurrence of these specif{}ied events.}
\promptnote{Inserted after the general rules for families whose events need an explicit boundary (here \texttt{cl\_007}, the pouring example in \cref{fig:annotation_boundaries}b). Under the swap, the A/B references in the scope are exchanged as well.}
\end{promptbox}
\vspace{2pt}
\begin{promptbox}{(d) Occurrence query (diagnostic, once per action)}
Use only the supplied chronological sampled images. Neither event is guaranteed to occur. Does the evidence establish occurrence of the TARGET event within this segment? ESTABLISHED means suf{}f{}icient visual evidence of the described occurrence or a diagnostic observed state change. NOT\_ESTABLISHED means occurrence is not established, not that it def{}initely did not happen. Objects, nearby hands, typical routines, or later possible actions alone are insuf{}f{}icient.\par Event A: \slot{Turn on the table lamp with the beige shade.}\par Event B: \slot{Open the black-covered notebook.}\par \slot{TARGET: Event A.}\par Return exactly ESTABLISHED or NOT\_ESTABLISHED.
\promptnote{Same images as the order query; asked for six instruction models and two Thinking configurations (\cref{app:occurrence}).}
\end{promptbox}
\end{minipage}
\caption{\textbf{Prompts of the main evaluation}, reproduced verbatim from the frozen request files. Frames are passed before the text as separate images in chronological order, inside each model\textquotesingle s native chat template; no system prompt is added.}
\label{fig:prompts_main}
\end{figure*}

\begin{figure*}[!t]
\centering\footnotesize
\begin{minipage}[t]{0.495\textwidth}\vspace{0pt}
\begin{promptbox}{(a) Variant: concise}
Judge the strict order of the two described events from these chronological sampled images only. The events are not guaranteed to occur. Answer A for A before B, B for B before A, or UNCLEAR if their occurrences or strict order are not established. Do not use expected routines.\par Event A: \slot{Turn on the table lamp with the beige shade.}\par Event B: \slot{Open the black-covered notebook.}\par\smallskip Return exactly one allowed label and no other text.
\end{promptbox}
\vspace{2pt}
\begin{promptbox}{(b) Variant: explicit evidence instruction}
Use only these chronological sampled images. First distinguish evidence that each described event occurred from evidence of their strict order. Even if both events occurred, their order may remain unknown. Never infer order from routine, object position alone, or missing observations. A means A before B; B means B before A. Return UNCLEAR unless both occurrences and their strict order are established.\par Event A: \slot{Turn on the table lamp with the beige shade.}\par Event B: \slot{Open the black-covered notebook.}\par\smallskip Return exactly one allowed label and no other text.
\end{promptbox}
\vspace{2pt}
\begin{promptbox}{(c) Variant: direct}
From the sampled images in chronological order, determine whether event A precedes event B, event B precedes event A, or the evidence cannot establish a strict order. Neither occurrence is assumed. Use A, B, or UNCLEAR respectively; do not f{}ill missing evidence using typical routines.\par Event A: \slot{Turn on the table lamp with the beige shade.}\par Event B: \slot{Open the black-covered notebook.}\par\smallskip Return exactly one allowed label and no other text.
\promptnote{Variants (a)--(c) and their majority vote are evaluated for four models (\cref{app:strategies}).}
\end{promptbox}
\end{minipage}\hfill
\begin{minipage}[t]{0.495\textwidth}\vspace{0pt}
\begin{promptbox}{(d) Staged checking: suffix appended to the main prompt}
Separate model predictions (fallible, not ground truth): \slot{A occurrence=ESTABLISHED}; \slot{B occurrence=ESTABLISHED}. Check these predictions against the same images. They do not determine event order and may be wrong. Return exactly A, B, or UNCLEAR.
\promptnote{The injected values are the model\textquotesingle s own answers to the two occurrence queries (or INVALID\_OUTPUT), never reference labels.}
\end{promptbox}
\vspace{2pt}
\begin{promptbox}{(e) Forced-choice text prior (no images, no UNCLEAR)}
No images or video are provided. Based only on the two descriptions, which event is more likely to happen f{}irst? Answer A if Event A happens f{}irst or B if Event B happens f{}irst; you must choose one.\par Event A: \slot{Turn on the table lamp with the beige shade.}\par Event B: \slot{Open the black-covered notebook.}\par Return exactly A or B.
\promptnote{Measures which order a model prefers from the descriptions alone; compare the committed answers in \cref{app:committed}.}
\end{promptbox}
\end{minipage}
\caption{\textbf{Prompt variants and controls}, verbatim, for the same family. Colored text is substituted per family.}
\label{fig:prompts_variants}
\end{figure*}

\Cref{fig:request} follows one query from frames to score. The frames of an input are passed as separate images in temporal order, followed by the prompt text, inside the model's native chat template; no system prompt is added and set names never appear. \Cref{fig:prompts_main} gives the prompts of the main evaluation verbatim, and \cref{fig:prompts_variants} the prompt variants and controls. Only the colored text changes between families; the swapped query exchanges the two descriptions and, where present, the A/B references in the scope sentence.

\subsection{Answer Extraction}\label{app:parsing}
One fixed, label-blind rule extracts answers for every model and every run (\cref{fig:request}). For Thinking configurations, only the text after the closing \texttt{</think>} counts, and an output without it is invalid. The final text must be exactly one allowed label (A, B, or UNCLEAR) or one of four aliases: ``A before B'' and ``A precedes B'' map to A, ``B before A'' and ``B precedes A'' to B. A label is also accepted when it is only wrapped in formatting, as in \texttt{\textbackslash boxed\{A\}}, \texttt{**A**}, ``A.'', or ``Answer: A'', or when it forms the last line of the output on its own, as in a short justification followed by a line containing only the label. The rule never searches an explanation for an answer letter: two labels (``A;B''), repeated labels (``AA''), free text without a final label, and outputs cut off at the token limit are invalid and count as wrong.

The exact-label check and the aliases were fixed before scoring; the formatting and last-line forms were added afterwards, when MiMo-VL in Thinking mode wrote nearly every answer as \texttt{\textbackslash boxed\{A\}}, and they were fixed before any accuracy was computed with them. They change no score of the configurations evaluated before them: apart from truncation, the only invalid outputs of those runs are two repeated labels from Gemma~4, which stay invalid. GLM-4.6V-Flash writes its final answer inside its native answer tokens (\texttt{<|begin\_of\_box|>}A\texttt{<|end\_of\_box|>}); when the output is exactly this wrapper, the wrapper is removed before the same rule, as the text before \texttt{</think>} is for Thinking configurations. This step was fixed from the answer formats of its 24 check queries, with the labels masked, before any accuracy was computed; no other model emits these tokens. For Gemma~4, the aliases recover 233 of 238 outputs that are not an exact label (59.2\% exact-label accuracy, 62.3\% extracted), with both correct and incorrect labels among them. Elsewhere the two forms recover 229 occurrence answers, 217 of them final lines that Qwen3.8-27B with Thinking writes after a short justification, two answers to a prompt variant, and 1,272 answers of the text-only priors: 996 boxed answers of MiMo-VL-7B-RL with Thinking and 276 forced choices that other models justify before giving them.

\section{Additional Results}\label{app:robustness}
\begin{table*}[!t]
\centering\scriptsize\setlength{\tabcolsep}{3.5pt}
\begin{tabular}{lrrrrrr}
\toprule
Model & Overall & Ord. & Insuf. & \PairHL & \PairKP & Family \\
\midrule
Gemma 4 12B & 62.3 [60.2, 64.5] & 56.6 [53.1, 60.2] & 68.3 [65.0, 71.5] & 33.5 [29.9, 37.2] & 36.3 [32.8, 39.9] & 20.0 [16.8, 23.4] \\
GLM-4.6V-Flash & 33.1 [31.7, 34.5] & 51.9 [51.2, 52.6] & 13.2 [10.6, 15.8] & 8.0 [6.3, 9.7] & 6.7 [5.3, 8.3] & 0.3 [0.0, 0.9] \\
InternVL3.5-8B & 43.7 [41.8, 45.6] & 73.5 [71.1, 76.0] & 12.2 [9.9, 14.7] & 8.7 [6.7, 11.0] & 11.0 [8.7, 13.5] & 7.3 [5.2, 9.4] \\
InternVL3.5-14B & 44.5 [42.6, 46.6] & 76.3 [73.7, 79.0] & 11.0 [8.5, 13.7] & 9.5 [7.2, 12.0] & 9.1 [6.9, 11.6] & 7.3 [5.3, 9.5] \\
Keye-VL-1.5-8B & 46.8 [44.8, 48.9] & 51.6 [48.5, 54.7] & 41.9 [38.3, 45.4] & 16.8 [14.3, 19.3] & 15.9 [13.5, 18.4] & 5.0 [3.3, 6.8] \\
LLaVA-OneVision-2-8B & 42.2 [40.6, 43.9] & 81.3 [78.8, 83.9] & 0.9 [0.3, 1.7] & 1.1 [0.4, 1.9] & 0.8 [0.2, 1.6] & 3.1 [1.8, 4.7] \\
MiMo-VL-7B-SFT & 44.5 [42.5, 46.5] & 61.4 [59.6, 63.2] & 26.6 [23.3, 30.0] & 19.1 [16.2, 22.0] & 15.8 [13.3, 18.5] & 5.9 [4.0, 7.9] \\
MiMo-VL-7B-RL & 54.5 [52.3, 56.6] & 67.3 [65.0, 69.6] & 40.9 [37.2, 44.5] & 28.1 [24.8, 31.4] & 27.0 [24.1, 30.0] & 10.7 [8.2, 13.3] \\
MiniCPM-V4.7 & 58.7 [56.7, 60.8] & 93.9 [92.1, 95.6] & 21.6 [18.3, 25.0] & 21.3 [17.9, 24.7] & 21.5 [18.1, 24.9] & 19.3 [16.1, 22.7] \\
Qwen3-VL-8B & 54.4 [52.1, 56.7] & 80.9 [78.0, 83.6] & 26.6 [23.1, 30.1] & 23.7 [20.5, 27.0] & 21.1 [18.0, 24.3] & 16.8 [13.7, 19.9] \\
Qwen3-VL-32B & 61.5 [59.2, 63.7] & 88.6 [86.3, 90.8] & 32.9 [29.2, 36.6] & 29.0 [25.4, 32.5] & 30.7 [27.0, 34.3] & 23.1 [19.7, 26.7] \\
Qwen3.8-27B & 73.4 [71.2, 75.5] & 77.0 [74.5, 79.6] & 69.5 [65.6, 73.2] & 52.2 [48.2, 56.1] & 52.6 [48.6, 56.4] & 33.0 [29.1, 36.9] \\
\midrule
MiMo-VL-7B-RL (Thinking) & 84.2 [82.5, 85.8] & 86.6 [84.3, 88.8] & 81.7 [79.1, 84.1] & 69.9 [66.4, 73.3] & 72.0 [68.7, 75.3] & 51.5 [47.3, 55.6] \\
Qwen3-VL-8B-Thinking & 72.6 [70.5, 74.7] & 86.7 [84.5, 88.7] & 57.8 [54.3, 61.2] & 51.3 [47.5, 54.9] & 50.3 [46.5, 54.1] & 32.1 [28.3, 36.0] \\
Qwen3-VL-32B-Thinking & 81.2 [79.3, 83.0] & 93.3 [91.5, 94.9] & 68.4 [64.9, 71.8] & 65.1 [61.2, 69.0] & 63.7 [60.0, 67.2] & 48.4 [43.9, 52.6] \\
Qwen3.8-27B (Thinking) & 89.3 [87.7, 90.8] & 95.0 [93.5, 96.3] & 83.3 [80.6, 86.0] & 80.1 [77.1, 83.0] & 80.4 [77.3, 83.4] & 67.5 [63.6, 71.4] \\
\midrule
Gemini 3.8 Flash$^\dagger$ & 94.4 [93.2, 95.5] & 92.0 [90.0, 93.8] & 96.9 [95.6, 98.1] & 88.8 [86.3, 91.2] & 90.3 [88.0, 92.5] & 81.3 [78.0, 84.7] \\
Gemini 4 Argon$^\dagger$ & 94.1 [92.9, 95.4] & 96.8 [95.6, 98.0] & 91.3 [89.1, 93.5] & 87.8 [85.3, 90.4] & 90.1 [87.8, 92.3] & 83.2 [80.2, 86.3] \\
GPT-6.1 Sol$^\dagger$ & 97.3 [96.5, 98.1] & 97.7 [96.6, 98.6] & 96.9 [95.6, 98.1] & 94.6 [92.8, 96.3] & 95.7 [94.1, 97.2] & 90.5 [88.0, 92.8] \\
\bottomrule
\end{tabular}

\caption{\textbf{Main metrics with 95\% cluster-bootstrap intervals} (\%). Rows are grouped as in \cref{tab:main}; $^\dagger$closed-source references. Intervals for the evidence types are in \cref{tab:uncertainty_strata}.}
\label{tab:intervals}
\end{table*}

\subsection{Main Metrics with Confidence Intervals}\label{app:intervals}
\Cref{tab:intervals} adds 95\% cluster-bootstrap intervals to the per-query and joint metrics of \cref{tab:main}, and \cref{tab:uncertainty_strata} to its evidence-type columns.

\begin{table}[!t]
\centering\footnotesize
\begin{tabular*}{\linewidth}{@{\extracolsep{\fill}}lrrr@{}}
\toprule
Model & PT & HO-Cap & CC4D \\
\midrule
Gemma 4 12B & 63.0 & 55.0 & 54.4 \\
GLM-4.6V-Flash & 33.6 & 26.2 & 26.5 \\
InternVL3.5-8B & 44.3 & 31.7 & 47.8 \\
InternVL3.5-14B & 44.2 & 46.2 & 50.7 \\
Keye-VL-1.5-8B & 47.7 & 32.9 & 44.9 \\
LLaVA-OneVision-2-8B & 42.1 & 38.3 & 52.9 \\
MiMo-VL-7B-SFT & 44.9 & 30.8 & 55.1 \\
MiMo-VL-7B-RL & 55.4 & 39.6 & 52.2 \\
MiniCPM-V4.7 & 59.5 & 48.3 & 52.9 \\
Qwen3-VL-8B & 54.5 & 53.8 & 53.7 \\
Qwen3-VL-32B & 62.7 & 43.8 & 54.4 \\
Qwen3.8-27B & 75.1 & 48.8 & 63.2 \\
\midrule
MiMo-VL-7B-RL (Thinking) & 84.6 & 85.0 & 71.3 \\
Qwen3-VL-8B-Thinking & 72.9 & 75.4 & 58.8 \\
Qwen3-VL-32B-Thinking & 82.4 & 67.9 & 64.7 \\
Qwen3.8-27B (Thinking) & 89.6 & 92.9 & 73.5 \\
\bottomrule
\end{tabular*}

\caption{\textbf{Overall accuracy by source dataset} (\%) for the open configurations: Perception Test (PT, 532 families), HO-Cap (30), and CaptainCook4D (CC4D, 17). Rows are grouped as in \cref{tab:main}.}
\label{tab:by_source}
\end{table}

\Cref{tab:by_source} splits overall accuracy by source dataset. Families from all three sources are screened, labeled, and reviewed under the same criteria (\cref{app:review}); HO-Cap and CaptainCook4D contribute few families, so their columns are less precise. In every source, each Thinking configuration is more accurate than its instruction counterpart.

\subsection{Evidence Types}\label{app:types}
\begin{figure*}[!t]
\centering
\includegraphics[width=\textwidth]{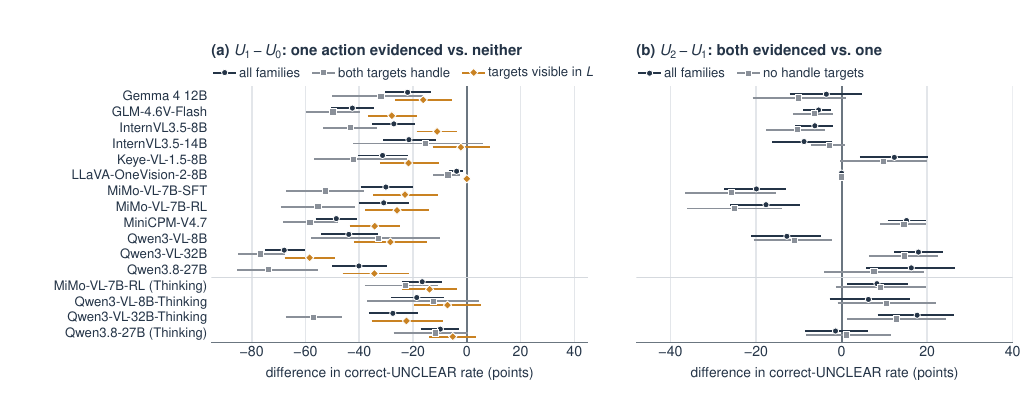}
\caption{\textbf{Differences between evidence types, with controls} (correct-UNCLEAR rate, points; 95\% unpaired cluster-bootstrap intervals). (a) $U_1-U_0$ over all families, among families whose two targets are both handle actions, and among $L$ inputs whose unevidenced targets are visible in at least one frame. (b) $U_2-U_1$ over all families and among families without handle targets.}
\label{fig:contrasts}
\end{figure*}

\begin{table*}[!t]
\centering\footnotesize
\begin{tabular}{lrrrrrr}
\toprule
Model & $U_0$ [95\% CI] & $U_1$ [95\% CI] & $U_2$ [95\% CI] & UNCLEAR on ord. & A/B cons. & A share \\
\midrule
Gemma 4 12B & 86.8 [81.7, 91.5] & 64.8 [57.6, 72.0] & 61.2 [56.4, 66.0] & 31.1 & 79.3 & 28.6 \\
GLM-4.6V-Flash & 48.2 [40.7, 55.6] & 5.6 [2.8, 9.2] & 0.2 [0.0, 0.7] & 0.1 & 7.3 & 92.4 \\
InternVL3.5-8B & 36.1 [29.3, 43.1] & 9.0 [5.1, 13.4] & 2.7 [1.4, 4.2] & 0.6 & 55.2 & 25.7 \\
InternVL3.5-14B & 31.9 [25.0, 39.3] & 10.4 [4.3, 17.7] & 1.6 [0.6, 2.8] & 0.7 & 70.5 & 57.3 \\
Keye-VL-1.5-8B & 58.8 [52.4, 65.2] & 27.5 [21.2, 34.2] & 39.8 [35.2, 44.3] & 25.1 & 52.7 & 47.8 \\
LLaVA-OneVision-2-8B & 3.8 [1.3, 6.8] & 0.0 [0.0, 0.0] & 0.0 [0.0, 0.0] & 0.0 & 77.7 & 46.5 \\
MiMo-VL-7B-SFT & 60.1 [52.8, 67.2] & 30.0 [23.6, 37.0] & 10.1 [7.2, 13.2] & 3.7 & 26.3 & 77.3 \\
MiMo-VL-7B-RL & 73.8 [67.4, 80.0] & 42.9 [35.7, 50.5] & 25.2 [21.2, 29.6] & 9.0 & 46.2 & 59.4 \\
MiniCPM-V4.7 & 50.0 [42.4, 57.4] & 1.5 [0.2, 3.5] & 16.6 [12.7, 20.9] & 0.2 & 87.8 & 48.9 \\
Qwen3-VL-8B & 66.6 [59.0, 73.6] & 22.7 [15.2, 30.8] & 9.9 [7.1, 12.9] & 1.5 & 81.0 & 39.4 \\
Qwen3-VL-32B & 74.4 [67.9, 80.6] & 6.5 [3.1, 10.5] & 24.4 [20.1, 28.7] & 1.2 & 83.9 & 37.7 \\
Qwen3.8-27B & 91.0 [86.5, 95.0] & 50.8 [41.8, 60.2] & 67.1 [62.2, 71.8] & 12.1 & 74.5 & 38.9 \\
\midrule
MiMo-VL-7B-RL (Thinking) & 89.7 [85.5, 93.5] & 73.1 [66.6, 79.4] & 81.4 [78.2, 84.4] & 10.8 & 85.9 & 28.5 \\
Qwen3-VL-8B-Thinking & 68.4 [62.0, 74.3] & 49.8 [41.9, 58.0] & 56.1 [51.2, 60.8] & 5.3 & 76.9 & 33.5 \\
Qwen3-VL-32B-Thinking & 79.6 [74.0, 84.9] & 52.1 [44.6, 60.1] & 69.8 [65.2, 74.3] & 3.3 & 88.9 & 32.3 \\
Qwen3.8-27B (Thinking) & 91.5 [87.6, 95.0] & 81.7 [75.2, 87.4] & 80.2 [76.3, 84.2] & 4.0 & 92.6 & 28.9 \\
\midrule
Gemini 3.8 Flash$^\dagger$ & 99.8 [99.4, 100.0] & 96.0 [92.7, 98.7] & 95.9 [93.9, 97.8] & 7.9 & 96.4 & 24.5 \\
Gemini 4 Argon$^\dagger$ & 98.2 [96.5, 99.6] & 91.5 [86.6, 95.6] & 88.1 [84.5, 91.6] & 2.8 & 96.6 & 26.7 \\
GPT-6.1 Sol$^\dagger$ & 99.5 [98.2, 100.0] & 96.2 [93.4, 98.6] & 96.1 [93.9, 97.9] & 2.3 & 98.4 & 25.8 \\
\bottomrule
\end{tabular}

\caption{\textbf{Correct-UNCLEAR rate by evidence type} (\%, 95\% intervals), share of orderable queries answered UNCLEAR, semantic A/B consistency over the 2,316 inputs, and share of A answers; $^\dagger$closed-source references. $U_0$/$U_1$/$U_2$ contain 554/480/1,220 queries from 143/124/301 families.}
\label{tab:uncertainty_strata}
\end{table*}

\begin{table*}[!t]
\centering\scriptsize\setlength{\tabcolsep}{3.5pt}
\resizebox{\textwidth}{!}{\begin{tabular}{lrrrrrr}
\toprule
Model & $U_1-U_0$ & $U_2-U_0$ & $U_1-U_0$ (H2) & $U_1-U_0$ (vis.) & $U_2-U_1$ & $U_2-U_1$ (H0) \\
\midrule
Gemma 4 12B & $-$22.0 [$-$30.4, $-$13.3] & $-$25.6 [$-$32.3, $-$18.5] & $-$32.0 [$-$50.1, $-$16.1] & $-$16.2 [$-$26.8, $-$5.6] & $-$3.6 [$-$12.1, 4.7] & $-$10.1 [$-$20.8, 1.0] \\
GLM-4.6V-Flash & $-$42.6 [$-$50.5, $-$34.5] & $-$47.9 [$-$55.4, $-$40.5] & $-$49.8 [$-$60.0, $-$39.6] & $-$27.9 [$-$36.8, $-$18.7] & $-$5.4 [$-$8.9, $-$2.5] & $-$6.3 [$-$11.4, $-$2.1] \\
InternVL3.5-8B & $-$27.1 [$-$35.2, $-$19.1] & $-$33.4 [$-$40.6, $-$26.5] & $-$43.4 [$-$53.3, $-$33.3] & $-$11.1 [$-$18.7, $-$3.6] & $-$6.3 [$-$10.8, $-$2.1] & $-$10.3 [$-$17.6, $-$3.8] \\
InternVL3.5-14B & $-$21.5 [$-$31.1, $-$11.5] & $-$30.3 [$-$37.8, $-$23.2] & $-$15.3 [$-$42.3, 6.0] & $-$2.2 [$-$12.7, 8.7] & $-$8.8 [$-$16.3, $-$2.3] & $-$2.8 [$-$7.2, 0.8] \\
Keye-VL-1.5-8B & $-$31.3 [$-$40.4, $-$21.9] & $-$19.0 [$-$26.7, $-$11.1] & $-$42.2 [$-$56.7, $-$22.4] & $-$21.7 [$-$32.3, $-$10.4] & 12.3 [4.3, 20.3] & 9.8 [$-$0.4, 20.1] \\
LLaVA-OneVision-2-8B & $-$3.8 [$-$6.8, $-$1.3] & $-$3.8 [$-$6.8, $-$1.3] & $-$7.0 [$-$12.4, $-$2.5] & 0.0 [0.0, 0.0] & 0.0 [0.0, 0.0] & 0.0 [0.0, 0.0] \\
MiMo-VL-7B-SFT & $-$30.1 [$-$39.5, $-$20.1] & $-$50.0 [$-$57.7, $-$42.2] & $-$52.6 [$-$67.2, $-$38.1] & $-$23.0 [$-$34.8, $-$10.8] & $-$19.9 [$-$27.5, $-$12.9] & $-$25.7 [$-$36.5, $-$15.4] \\
MiMo-VL-7B-RL & $-$30.9 [$-$40.2, $-$21.5] & $-$48.6 [$-$56.1, $-$40.7] & $-$55.4 [$-$69.2, $-$41.4] & $-$26.0 [$-$37.7, $-$14.2] & $-$17.7 [$-$26.0, $-$9.7] & $-$25.0 [$-$36.0, $-$13.9] \\
MiniCPM-V4.7 & $-$48.5 [$-$56.2, $-$40.8] & $-$33.4 [$-$41.7, $-$25.0] & $-$58.3 [$-$68.4, $-$48.0] & $-$34.3 [$-$43.6, $-$24.9] & 15.2 [10.8, 19.7] & 14.6 [9.0, 19.7] \\
Qwen3-VL-8B & $-$43.9 [$-$54.2, $-$33.1] & $-$56.7 [$-$64.1, $-$48.6] & $-$32.8 [$-$57.8, $-$10.1] & $-$28.4 [$-$41.8, $-$14.8] & $-$12.8 [$-$21.1, $-$4.8] & $-$11.1 [$-$20.6, $-$2.2] \\
Qwen3-VL-32B & $-$67.9 [$-$75.1, $-$60.3] & $-$49.9 [$-$57.7, $-$42.1] & $-$76.7 [$-$85.1, $-$67.8] & $-$58.5 [$-$67.8, $-$49.1] & 18.0 [12.3, 23.6] & 14.8 [6.4, 22.4] \\
Qwen3.8-27B & $-$40.1 [$-$50.0, $-$29.8] & $-$23.8 [$-$30.1, $-$17.2] & $-$73.7 [$-$85.3, $-$55.4] & $-$34.4 [$-$46.2, $-$21.7] & 16.3 [5.8, 26.4] & 7.6 [$-$4.0, 19.2] \\
\midrule
MiMo-VL-7B-RL (Thinking) & $-$16.6 [$-$24.2, $-$9.2] & $-$8.3 [$-$13.3, $-$3.1] & $-$22.8 [$-$37.9, $-$10.7] & $-$13.8 [$-$24.0, $-$3.5] & 8.3 [1.2, 15.6] & 9.0 [$-$1.3, 19.7] \\
Qwen3-VL-8B-Thinking & $-$18.6 [$-$28.2, $-$8.5] & $-$12.3 [$-$19.8, $-$4.4] & $-$12.5 [$-$37.0, 4.4] & $-$7.2 [$-$19.6, 5.4] & 6.3 [$-$2.8, 15.9] & 10.5 [$-$0.9, 22.0] \\
Qwen3-VL-32B-Thinking & $-$27.5 [$-$36.4, $-$18.1] & $-$9.8 [$-$16.8, $-$2.7] & $-$57.0 [$-$67.3, $-$46.3] & $-$22.5 [$-$35.4, $-$9.0] & 17.7 [8.4, 26.2] & 12.8 [1.4, 24.3] \\
Qwen3.8-27B (Thinking) & $-$9.8 [$-$17.2, $-$2.9] & $-$11.3 [$-$16.5, $-$5.9] & $-$11.6 [$-$27.2, 0.4] & $-$5.2 [$-$14.0, 3.3] & $-$1.4 [$-$8.6, 6.1] & 1.1 [$-$8.4, 11.5] \\
\bottomrule
\end{tabular}
}
\caption{\textbf{Differences between evidence types} (correct-UNCLEAR rate, points; 95\% unpaired cluster-bootstrap intervals), as plotted in \cref{fig:contrasts}. H2: families whose two targets are both handle actions; vis.: $L$ inputs whose unevidenced targets are visible; H0: families without handle targets.}
\label{tab:type_differences}
\end{table*}

\Cref{tab:uncertainty_strata} gives the correct-UNCLEAR rate of each type with intervals, and \cref{fig:contrasts} and \cref{tab:type_differences} the differences between types. Both $U_1-U_0$ and $U_2-U_0$ have intervals below zero for every configuration, the smallest being $U_1-U_0$ of Qwen3.8-27B with Thinking ($-$9.8, interval $-$17.2 to $-$2.9); $U_2-U_1$ is significantly positive for Keye-VL-1.5-8B, Qwen3-VL-32B, MiniCPM-V4.7, Qwen3.8-27B, Qwen3-VL-32B-Thinking, and MiMo-VL-7B-RL with Thinking (8.3--18.0 points) and significantly negative for GLM-4.6V-Flash, both InternVL3.5 models, MiMo-VL-7B-SFT and -RL, and Qwen3-VL-8B (5.4--19.9 points). All but one of the 143 $U_0$ families come from Perception Test; restricted to Perception Test families, $U_1-U_0$ and $U_2-U_0$ stay below zero for every configuration. Because types mostly differ between families, these differences largely compare different events and use an unpaired cluster bootstrap; \cref{app:matched} compares types within families. The table also gives the share of orderable queries answered UNCLEAR (3.3--10.8\% for the four Thinking configurations, 25.1\% for Keye-VL-1.5-8B, 31.1\% for Gemma~4). A/B consistency and the share of A answers expose letter preferences, most strongly for GLM-4.6V-Flash (92.4\% A answers) and MiMo-VL-7B-SFT; consistency alone does not imply correctness, since always answering UNCLEAR is perfectly consistent.

\subsection{Matched Diagnostic Subset}\label{app:matched}
\begin{table*}[!t]
\centering\scriptsize\setlength{\tabcolsep}{3pt}
\begin{tabular}{lrrrrrrrrrr}
\toprule
Model & $U_0$ & $U_1$ & $U_2$ & $U_1-U_0$ & $U_2-U_1$ & Commit & Agree & Shown 1st & Res.~$U_1$ & Other $U_1$ \\
\midrule
Gemma 4 12B & 95.7 & 74.6 & 66.0 & $-$21.0 [$-$24.1, $-$18.1] & $-$14.2 [$-$19.2, $-$8.9] & 22.5 & 63.3 & 43.3 & 75.2 & 74.6 \\
GLM-4.6V-Flash & 63.7 & 5.1 & 0.4 & $-$59.1 [$-$63.3, $-$54.8] & $-$5.6 [$-$7.3, $-$3.9] & 93.0 & 51.3 & 50.9 & 0.2 & 7.5 \\
InternVL3.5-8B & 40.9 & 3.6 & 1.3 & $-$37.5 [$-$41.8, $-$33.3] & $-$3.0 [$-$4.8, $-$1.4] & 92.8 & 60.8 & 45.9 & 0.8 & 4.8 \\
InternVL3.5-14B & 43.5 & 6.5 & 2.2 & $-$37.1 [$-$41.6, $-$32.7] & $-$4.0 [$-$6.0, $-$2.0] & 87.3 & 62.6 & 44.7 & 0.6 & 9.7 \\
Keye-VL-1.5-8B & 74.9 & 37.6 & 38.0 & $-$36.8 [$-$40.7, $-$32.9] & $-$5.1 [$-$9.1, $-$1.3] & 52.8 & 57.1 & 53.3 & 38.0 & 37.9 \\
LLaVA-OneVision-2-8B & 4.6 & 0.0 & 0.0 & $-$4.6 [$-$6.6, $-$2.9] & +0.0 [0.0, 0.0] & 100.0 & 61.9 & 41.5 & 0.0 & 0.0 \\
MiMo-VL-7B-SFT & 84.8 & 36.9 & 10.9 & $-$48.2 [$-$51.8, $-$44.5] & $-$25.3 [$-$29.0, $-$21.4] & 58.2 & 56.1 & 48.4 & 14.5 & 47.8 \\
MiMo-VL-7B-RL & 89.4 & 51.5 & 25.3 & $-$37.8 [$-$41.2, $-$34.4] & $-$25.4 [$-$29.8, $-$20.7] & 44.2 & 60.7 & 41.7 & 25.6 & 63.9 \\
MiniCPM-V4.7 & 74.0 & 6.6 & 16.4 & $-$67.5 [$-$71.7, $-$63.4] & +9.2 [4.9, 13.7] & 91.2 & 74.2 & 39.3 & 0.6 & 9.7 \\
Qwen3-VL-8B & 84.7 & 11.7 & 9.8 & $-$73.2 [$-$76.7, $-$69.5] & $-$2.6 [$-$6.2, 1.3] & 86.6 & 70.2 & 50.4 & 4.3 & 15.9 \\
Qwen3-VL-32B & 86.4 & 10.8 & 23.5 & $-$75.4 [$-$78.7, $-$72.0] & +11.0 [6.3, 15.7] & 87.5 & 69.3 & 57.1 & 6.2 & 13.1 \\
Qwen3.8-27B & 96.7 & 60.2 & 70.5 & $-$36.0 [$-$39.5, $-$32.6] & +8.9 [3.6, 14.1] & 38.3 & 69.2 & 46.2 & 42.8 & 68.9 \\
\midrule
MiMo-VL-7B-RL (Thinking) & 96.0 & 76.6 & 81.6 & $-$19.2 [$-$21.8, $-$16.6] & +6.1 [1.4, 10.8] & 22.0 & 86.6 & 13.0 & 57.2 & 86.4 \\
Qwen3-VL-8B-Thinking & 80.5 & 45.4 & 57.0 & $-$35.4 [$-$39.5, $-$31.2] & +18.0 [12.3, 23.6] & 51.8 & 80.5 & 19.0 & 11.2 & 62.0 \\
Qwen3-VL-32B-Thinking & 90.5 & 46.2 & 71.2 & $-$44.4 [$-$47.9, $-$40.9] & +31.0 [25.2, 36.7] & 51.5 & 79.6 & 28.3 & 16.5 & 61.4 \\
Qwen3.8-27B (Thinking) & 96.5 & 61.9 & 82.0 & $-$34.6 [$-$37.8, $-$31.5] & +27.8 [22.9, 32.8] & 37.1 & 95.2 & 3.5 & 24.0 & 80.7 \\
\midrule
Gemini 3.8 Flash$^\dagger$ & 100.0 & 96.1 & 96.4 & $-$3.6 [$-$4.9, $-$2.4] & +0.4 [$-$2.5, 3.1] & 3.5 & 98.2 & 0.0 & 94.0 & 97.4 \\
Gemini 4 Argon$^\dagger$ & 99.3 & 89.3 & 87.8 & $-$9.3 [$-$11.5, $-$7.2] & $-$2.7 [$-$6.9, 1.7] & 9.1 & 98.3 & 0.0 & 85.5 & 91.8 \\
GPT-6.1 Sol$^\dagger$ & 99.9 & 91.1 & 97.3 & $-$8.4 [$-$10.5, $-$6.5] & +7.7 [4.8, 10.6] & 8.4 & 99.3 & 0.0 & 80.4 & 96.5 \\
\bottomrule
\end{tabular}

\caption{\textbf{Matched subset} (\%; differences in points with 95\% cluster-bootstrap intervals). $U_0$/$U_1$/$U_2$: correct-UNCLEAR rate on the subset's four-frame inputs of each type, including each family's $L$. $U_1-U_0$, $U_2-U_1$: differences within families. Commit: $U_1$ queries answered with an order where the same family's $U_0$ inputs were answered UNCLEAR. Agree: committed $U_1$ answers that match the order shown by $K$. Shown 1st: committed $U_1$ answers that put the shown action first when $K$ shows it to be the second. Res.~$U_1$: correct-UNCLEAR rate on the 258 $U_1$ inputs whose first frame already shows the result of the unestablished action; Other $U_1$: on the other 513. $^\dagger$Closed-source references.}
\label{tab:matched}
\end{table*}

\begin{table*}[!t]
\centering\scriptsize\setlength{\tabcolsep}{6pt}
\begin{tabular}{lrrrr}
\toprule
Thinking pair & $U_1$ withholding & $U_1-U_0$ & $U_2-U_1$ & $U_2-U_0$ \\
\midrule
MiMo-VL-7B-RL$\to$Thinking & +25.1 [22.0, 28.3] & +18.6 [15.0, 22.1] & +31.5 [25.1, 37.9] & +54.2 [47.7, 60.5] \\
Qwen3-VL-8B$\to$Thinking & +33.7 [30.4, 37.2] & +37.8 [33.2, 42.3] & +20.6 [14.4, 26.7] & +54.4 [47.7, 61.0] \\
Qwen3-VL-32B$\to$Thinking & +35.4 [32.2, 38.6] & +31.0 [26.9, 35.2] & +20.0 [13.6, 26.4] & +46.4 [39.7, 52.9] \\
Qwen3.8-27B$\to$Thinking & +1.7 [$-$2.1, 5.4] & +1.4 [$-$2.8, 5.4] & +19.0 [13.1, 24.9] & +11.6 [6.3, 16.9] \\
\bottomrule
\end{tabular}

\caption{\textbf{Thinking on the matched subset} (points; Thinking minus its instruction counterpart, paired by family and assignment; 95\% cluster-bootstrap intervals). $U_1$ withholding: change in the correct-UNCLEAR rate on $U_1$; other columns: change in the within-family difference.}
\label{tab:matched_thinking}
\end{table*}

\begin{figure}[t]
\centering
\includegraphics[width=\columnwidth]{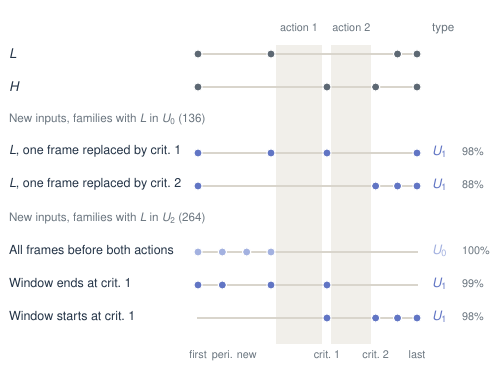}
\caption{\textbf{Construction of the matched subset} (schematic, not to scale). Dots mark the four frames of each input on the family's time axis; crit.~1 and crit.~2 are the critical frames of $H$, peri.\ the peripheral frame of $P$ near the start, and new a frame extracted for this input. Type: the type that most inputs of a construction receive, with its share; every input is labeled from its own frames.}
\label{fig:matched}
\end{figure}
Because evidence types mostly differ between families, the matched diagnostic subset adds new four-frame inputs to 400 families with orderable $H$ and $K$, so that inputs of different types share an episode and a question. Each new input takes four frames of the family in temporal order and changes which actions they evidence (\cref{fig:matched}). In the 136 families whose $L$ was in $U_0$ when the subset was built, one frame of $L$ is replaced by a critical frame of $H$, which keeps the window and frame count of $L$ but evidences one action. In the 264 families whose $L$ was in $U_2$, the four frames either precede both changes, using one newly extracted frame with the family's preprocessing, or form a window that ends at the first critical frame or starts there, so that one of the two changes falls outside the window. The new inputs reuse the family's two prompts verbatim, so only the images differ from the benchmark queries; every input is queried under both assignments by all sixteen open configurations and the three references and scored with the rule of \cref{app:parsing}. Labels follow the criteria of \cref{app:annotation}; they were drafted with Claude, and one annotator reviewed all of them, against at least two for every benchmark item (\cref{app:review}). Together with each family's $L$, the subset contains 388 $U_0$, 778 $U_1$, 276 $U_2$, and 11 orderable four-frame inputs; 388 families come from Perception Test, 11 from CaptainCook4D, and one from HO-Cap. The subset does not enter the main metrics.

\Cref{tab:matched} and \cref{fig:types}b compare types within families: for each family and assignment, the correct-UNCLEAR rate of a type is averaged over the family's inputs of that type, and differences are averaged over the 760 (family, assignment) units with both $U_0$ and $U_1$ inputs and the 546 with both $U_1$ and $U_2$. Every configuration withholds less often on $U_1$ than on $U_0$ of the same family: the open models by 19--75 points, except LLaVA-OneVision-2-8B ($-$4.6), which almost never withholds, and the references by 4--9. This includes Qwen3.8-27B with Thinking, whose benchmark difference is the smallest. Across the sixteen open configurations, the within-family $U_1-U_0$ follows the benchmark's (Pearson $r=0.87$). For $U_2-U_1$, whose sign differs between models, the subset agrees in sign with the benchmark for eleven of the twelve configurations whose benchmark difference is significant, ten of them significantly; Keye-VL-1.5-8B is the exception. This agreement depends more on which $U_1$ inputs are included: without the 258 $U_1$ inputs discussed below, the sign agrees for nine of the twelve configurations, seven of them significantly.

Evidence of one action turns many correct withholdings into answers. Where a model answered the $U_0$ inputs of a family UNCLEAR, the instruction models answer 22--100\% of the same family's $U_1$ inputs with an order (Commit; 86.6--100\% for seven of the twelve), the Thinking configurations 22--52\%, and the references 4--9\%. The committed $U_1$ answers of the instruction models match the order shown by $K$ in only 51--74\% of cases (Agree): they put the shown action first in 53--97\% of cases when $K$ shows it to be the first action, but also in 39--57\% when $K$ shows it to be the second (Shown 1st). The committed answers of the Thinking configurations match the true order in 80--95\% of cases, and those of the references in 98--99\%.

In 258 $U_1$ inputs, the window starts right after the first action, so the first frame already shows the result of an action that no frame establishes (Res.~$U_1$). The order query asks which action occurs first in the recorded segment (\cref{fig:prompts_main}a), and an end state without the state before it does not show that the action took place within the segment (\cref{app:annotation}): the object may have been in that state all along. The references withhold on 80--94\% of these queries. Thirteen of the sixteen open configurations withhold less often on these inputs than on the other $U_1$ inputs, including all four Thinking configurations (11--57\% vs.\ 61--86\%). A committed answer of Qwen3.8-27B with Thinking, drawn at random among these inputs (seed 20261106), shows the inference. Asked whether the green or the red triangular piece is first placed on the table, it reasons: ``In frame 1, green is already on the table and red is not. \ldots\ If red had been placed first, red would be present in frame 1. It's not. So green must have been placed before red.'' This accounts for the Thinking pairs in \cref{tab:matched_thinking}. Thinking raises $U_1$ withholding by 25--35 points for Qwen3-VL-8B, Qwen3-VL-32B, and MiMo-VL-7B-RL, but not for Qwen3.8-27B (+1.7, interval $-$2.1 to 5.4), whose withholding on these inputs falls from 42.8\% to 24.0\% with Thinking while it rises from 68.9\% to 80.7\% on the other $U_1$ inputs. In all four pairs, Thinking raises withholding on $U_2$ more than on $U_1$, by 19--32 points.

The within-family comparison does not depend on these inputs. Without them, every configuration still withholds less often on $U_1$ than on $U_0$ of the same family, with every 95\% interval below zero: the open models by 8--71 points, except LLaVA-OneVision-2-8B ($-$4.6), and the references by 2--6. The within-family differences then follow the benchmark's even more closely (Pearson $r=0.89$), and Thinking raises $U_1$ withholding in all four pairs: by 45.1, 46.3, and 21.5 points for Qwen3-VL-8B, Qwen3-VL-32B, and MiMo-VL-7B-RL, and by 8.7 points for Qwen3.8-27B (interval 4.1 to 13.3).

\subsection{Controls for Action Class and Visibility}\label{app:controls}
The matched subset compares evidence types within families (\cref{app:matched}); the controls here stratify the benchmark's own inputs instead. Handle actions usually leave no lasting change and are the most common class in families whose insufficient inputs miss an action (\cref{fig:composition}b). We therefore fix the number of handle targets (\cref{fig:contrasts}): $U_1-U_0$ among families whose two targets are both handle actions (79 $U_0$ and 30 $U_1$ families), and $U_2-U_1$ among families without handle targets (61 $U_1$ and 294 $U_2$ families). The drop from $U_0$ to $U_1$ remains for thirteen of the sixteen configurations, by 7--77 points. The excess of $U_2$ over $U_1$ withholding remains for Qwen3-VL-32B, MiniCPM-V4.7, and Qwen3-VL-32B-Thinking, which withhold on 10.2\%, 2.5\%, and 58.2\% of $U_1$ inputs without handle targets, but not for Qwen3.8-27B, Keye-VL-1.5-8B, or MiMo-VL-7B-RL with Thinking, whose $U_1$ withholding is 18.3\%, 22.1\%, and 69.2\% with two handle targets against 61.1\%, 30.3\%, and 73.0\% without. The Thinking gains also remain: relative to their instruction counterparts, the four Thinking configurations raise correct withholding by 24--64 points on $U_1$ with two handle targets, by 18--48 on $U_1$ without handle targets, and by 12--56 on $U_2$ without handle targets, while $U_0$ with two handle targets rises by at most 11 points.

A target object that never enters the frames is a natural way for an action to be missing from the frames. For every insufficient $L$ input of type $U_0$ or $U_1$, AI assistants recorded whether the main object of each unevidenced target appears in at least one frame, and we spot-checked the marks; these are analysis metadata, not labels. In 28 of the 143 families whose $L$ input is in $U_0$, at least one target is off-screen, mostly hold-up actions whose objects are brought in from outside the view; in 98 both are visible, and 17 are uncertain. The unevidenced target is visible in 103 of the 113 families whose $L$ input is in $U_1$. With an off-screen target, fourteen configurations answer UNCLEAR on 88--96\% of these inputs. Restricted to visible targets, $U_1$ withholding stays below $U_0$ for twelve of the sixteen configurations, by smaller margins; the exceptions are InternVL3.5-14B, Qwen3-VL-8B-Thinking, Qwen3.8-27B with Thinking, and LLaVA-OneVision-2-8B, which withholds on almost none of either type.

\subsection{Occurrence Queries}\label{app:occurrence}
\begin{table*}[!t]
\centering\scriptsize\setlength{\tabcolsep}{4pt}
\begin{tabular}{lrrrrrrrrrr}
\toprule
Model & Occ. acc. & Recall E & Recall NE & Bal. acc. & $U_1$ ordered & Est. first & $U_1$ queries & $U_2$ order err. & $U_2$ queries & $U_2$ families \\
\midrule
Gemma 4 12B & 60.2 & 53.5 & 92.9 & 73.2 & 43.4 & 59.0 & 242 & 62.5 & 368 & 107 \\
LLaVA-OneVision-2-8B & 78.9 & 79.9 & 74.1 & 77.0 & 100.0 & 63.1 & 214 & 100.0 & 732 & 193 \\
MiniCPM-V4.7 & 80.3 & 77.2 & 95.2 & 86.2 & 99.4 & 77.7 & 352 & 84.9 & 610 & 163 \\
Qwen3-VL-8B & 81.5 & 79.7 & 90.1 & 84.9 & 87.2 & 93.3 & 274 & 91.5 & 764 & 193 \\
Qwen3-VL-32B & 83.7 & 82.0 & 91.9 & 87.0 & 98.1 & 93.7 & 370 & 74.1 & 758 & 198 \\
Qwen3.8-27B & 34.6 & 21.2 & 99.7 & 60.5 & 68.6 & 89.7 & 156 & 33.3 & 6 & 2 \\
\midrule
Qwen3-VL-32B-Thinking & 90.3 & 92.2 & 81.0 & 86.6 & 41.9 & 76.7 & 358 & 26.1 & 1,004 & 260 \\
Qwen3.8-27B (Thinking) & 91.0 & 92.2 & 85.0 & 88.6 & 12.6 & 56.2 & 380 & 19.4 & 970 & 264 \\
\bottomrule
\end{tabular}

\caption{\textbf{Occurrence queries} (\%). Recall E/NE: recall on established and not established targets; Bal. acc.: their mean. Among $U_1$ queries whose two occurrence answers are both correct: the share answered with an order ($U_1$ ordered) and, of those, the share that puts the established action first (Est. first). $U_2$ order err.: order-error rate on $U_2$ queries whose two occurrence answers are both correct. Lower block: Thinking configurations.}
\label{tab:occurrence_full}
\end{table*}

Six instruction models (Gemma~4, LLaVA-OneVision-2, MiniCPM-V4.7, Qwen3-VL-8B and 32B, and Qwen3.8-27B) and two Thinking configurations (Qwen3-VL-32B-Thinking and Qwen3.8-27B with Thinking) also answer, for each input and target, whether the action is established, with the same images as the order queries (\cref{fig:prompts_main}d; 4,632 calls per model, with the output limits of the order queries). \Cref{tab:occurrence_full} reports occurrence accuracy, recall per class, and the order answers on $U_1$ and $U_2$ inputs whose two occurrence answers are both correct. The reference labels are imbalanced (3,838 established and 794 not established targets), so balanced accuracy is the more informative summary. The conditional subset depends on each model's own answers: Qwen3.8-27B answers NOT\_ESTABLISHED for most established targets, which leaves six $U_2$ queries from two families, so its $U_2$ rate is not interpreted. For the other five instruction models, the order answer on these inputs is wrong in 62--100\% of cases, so judging both occurrences correctly when asked separately does not lead them to withhold the order. On $U_1$ inputs where a model correctly judges that one action is established and the other is not, five of the six instruction models still order the actions in 68.6--100\% of cases (Gemma~4: 43.4\%), and these orders put the established action first in 63.1--93.7\%: correctly reporting the missing occurrence does not lead them to withhold either. The two answers come from separate queries and describe behavior across tasks.

With Thinking, what a model reports carries over to its order answer far more often. The two Thinking configurations judge occurrence about as accurately as the best instruction models (balanced accuracy 86.6\% and 88.6\%), and Qwen3.8-27B no longer answers NOT\_ESTABLISHED for most established targets (recall 92.2\% with Thinking against 21.2\% without). Where their two occurrence answers are correct, Qwen3-VL-32B-Thinking and Qwen3.8-27B with Thinking order $U_1$ inputs in 41.9\% and 12.6\% of cases, against 98.1\% and 68.6\% for the same models without Thinking, and fail to withhold on $U_2$ inputs in 26.1\% and 19.4\%, against 74.1\% for Qwen3-VL-32B. Four outputs of Qwen3-VL-32B-Thinking are truncated and six of Qwen3.8-27B with Thinking carry no final label; they count as wrong occurrence answers.

\subsection{Committed Answers and Text-Only Priors}\label{app:committed}

\begin{table*}[!t]
\centering\footnotesize
\begin{tabular}{lrrrrrr}
\toprule
Model & A share & Agree with $K$ & Content & True first & Content (frames) & True first (frames) \\
\midrule
Gemma 4 12B & 71.6 & 49.5 [47.2, 51.8] & 176 & 48.3 [41.0, 55.8] & 223 & 68.2 [60.1, 76.2] \\
GLM-4.6V-Flash$^*$ & 99.6 & 50.4 [50.1, 50.9] & 5 & 100.0 [100.0, 100.0] & 7 & 100.0 [100.0, 100.0] \\
InternVL3.5-8B & 80.6 & 51.9 [49.6, 54.2] & 178 & 56.2 [48.9, 63.6] & 436 & 63.8 [57.7, 69.7] \\
InternVL3.5-14B & 88.0 & 48.6 [46.8, 50.3] & 102 & 41.2 [31.7, 50.6] & 646 & 61.5 [56.6, 66.5] \\
Keye-VL-1.5-8B & 94.5 & 50.9 [49.6, 52.2] & 64 & 57.8 [45.8, 69.9] & 275 & 60.4 [52.5, 68.1] \\
LLaVA-OneVision-2-8B & 75.5 & 52.4 [49.8, 55.0] & 237 & 55.7 [49.6, 62.1] & 776 & 63.4 [58.9, 68.0] \\
MiMo-VL-7B-SFT & 10.0 & 50.7 [49.1, 52.4] & 93 & 54.8 [44.4, 65.1] & 40 & 82.5 [64.9, 97.4] \\
MiMo-VL-7B-RL & 11.1 & 51.2 [49.6, 52.8] & 85 & 57.6 [47.1, 68.2] & 145 & 69.7 [60.0, 79.0] \\
MiniCPM-V4.7 & 95.3 & 51.2 [50.0, 52.5] & 54 & 63.0 [50.0, 75.5] & 637 & 62.5 [57.3, 67.5] \\
Qwen3-VL-8B & 84.7 & 50.3 [48.3, 52.4] & 148 & 51.4 [43.5, 59.4] & 590 & 58.0 [52.5, 63.3] \\
Qwen3-VL-32B & 84.5 & 51.8 [49.7, 54.0] & 159 & 56.6 [49.0, 64.2] & 527 & 61.5 [55.8, 67.2] \\
Qwen3.8-27B & 48.7 & 50.9 [48.7, 53.1] & 168 & 53.0 [45.6, 60.5] & 142 & 76.1 [66.4, 86.0] \\
\midrule
MiMo-VL-7B-RL (Thinking) & 81.2 & 51.9 [49.3, 54.4] & 179 & 55.3 [48.0, 62.6] & 93 & 83.9 [74.4, 92.1] \\
Qwen3-VL-32B-Thinking & 64.6 & 51.9 [48.7, 55.0] & 317 & 53.6 [48.1, 59.0] & 238 & 82.8 [76.5, 88.8] \\
Qwen3.8-27B (Thinking) & 62.0 & 49.7 [46.6, 53.0] & 365 & 49.6 [44.6, 54.6] & 115 & 86.1 [78.1, 93.1] \\
\bottomrule
\end{tabular}

\caption{\textbf{Text-only priors and committed answers} (\%, 95\% cluster-bootstrap intervals). Without images (\cref{fig:prompts_variants}e): share of A answers, agreement with the order shown by the family's $K$ input, families with a content-based choice (the same action under both assignments), and the share of these that put the true first action first. Frames: the same for committed content-based answers on insufficient $L$ and $P$ inputs. $^*$GLM-4.6V-Flash gives the same letter under both assignments in all but five families without images and seven committed cases with frames, so its estimates are not interpreted. Qwen3-VL-8B-Thinking is not included: its 24-query check truncated seven outputs at 4,096 tokens.}
\label{tab:text_prior}
\end{table*}

When a model commits to an order on an insufficient input, we compare its answer with the order shown by the same family's $K$ input under the same assignment. Over complete pairs a fixed letter preference agrees half of the time, because the swap reverses the correct order; committed answers need not cover both assignments, and on them the best fixed letter agrees on 50.0--58.7\% per configuration and type. Agreement lies between 49.8\% and 90.9\%, and its interval lies above this letter baseline in 34 of the 48 configuration--type cells (\cref{fig:committed}). The two target descriptions alone do not reveal the order: forced to choose without any image (\cref{fig:prompts_variants}e), the fifteen open configurations we ran agree with the order shown by $K$ on 48.6--52.4\% of queries (\cref{tab:text_prior}). Two of them are treated separately: the estimates of GLM-4.6V-Flash, which gives the same letter under both assignments in nearly every family, are not interpreted, and Qwen3.8-27B with Thinking, which was run separately, is reported below. For each of the other thirteen, the 95\% interval includes 50\%, although most strongly prefer one letter (10--95\% A answers).

Letter preferences pull both measures toward 50\%, so we also keep only content-based choices, where a model picks the same action under both letter assignments. Without images such choices make up 9--60\% of the families a model answers with a letter under both assignments, and they put the true first action first in 41.2--63.0\% of them, with every interval including 50\%; on insufficient inputs, the committed content-based answers of the same thirteen configurations put it first in 58.0--83.9\%, with every interval above 50\%. On the same families, the frames raise this share by 12.3 points (95\% interval 5.0--19.5, pooled over the thirteen configurations; positive for eleven of them). Qwen3.8-27B with Thinking, run separately with the same requests, behaves alike: without images it agrees with the order shown by $K$ on 49.7\% of queries (interval 46.6--53.0) and its content-based choices put the true first action first in 49.6\% of families (44.6--54.6), whereas its committed content-based answers on insufficient inputs do so in 86.1\% (78.1--93.1). For Qwen3-VL-32B-Thinking and MiMo-VL-7B-RL with Thinking the contrast is sharp: without images their content-based choices put the true first action first in 53.6\% and 55.3\% of families, both intervals including 50\%, and with the frames of insufficient inputs in 82.8\% and 83.9\%.

Committed $U_1$ answers also follow a simple rule about the unseen action: they tend to put the evidenced action first (51.9--87.8\% of committed answers, with the 95\% interval above 50\% for eleven of the sixteen configurations and below it for none), although the evidenced action is the true first action in only 52.6\% of these inputs. Models thus tend to place the action that no frame shows second, which the prompt rules out. The matched subset separates the two cases: the committed $U_1$ answers of the instruction models put the shown action first in 53--97\% of cases when $K$ shows it to be first and still in 39--57\% when $K$ shows it to be second (\cref{app:matched}).

\subsection{Responses to Added Evidence}\label{app:added}
\begin{table*}[!t]
\centering\footnotesize
\begin{tabular}{lrrrrrrrr}
\toprule
Model & $H$ & $L$ & $K$ & $P$ & $L$ \& $K$ & $K$ (pairs) & $K\mid L$ correct & $L$ \& $P$ \\
\midrule
Gemma 4 12B & 51.1 & 71.4 & 62.7 & 64.0 & 42.4 & 62.7 & 58.2 & 61.1 \\
GLM-4.6V-Flash & 51.6 & 15.5 & 51.0 & 14.0 & 7.6 & 51.6 & 53.2 & 13.6 \\
InternVL3.5-8B & 71.7 & 13.2 & 73.1 & 16.8 & 8.9 & 73.7 & 82.4 & 12.6 \\
InternVL3.5-14B & 75.1 & 13.9 & 75.6 & 13.3 & 9.6 & 76.6 & 82.2 & 12.1 \\
Keye-VL-1.5-8B & 52.3 & 40.5 & 49.1 & 45.4 & 13.5 & 49.0 & 33.7 & 38.3 \\
LLaVA-OneVision-2-8B & 80.1 & 4.1 & 80.2 & 4.3 & 1.1 & 80.8 & 100.0 & 3.8 \\
MiMo-VL-7B-SFT & 63.6 & 30.4 & 58.3 & 25.6 & 18.2 & 59.0 & 61.9 & 24.5 \\
MiMo-VL-7B-RL & 68.5 & 42.5 & 64.8 & 42.1 & 27.7 & 65.1 & 66.7 & 37.9 \\
MiniCPM-V4.7 & 92.3 & 24.7 & 92.9 & 25.0 & 21.5 & 93.8 & 97.5 & 23.0 \\
Qwen3-VL-8B & 78.3 & 30.8 & 80.7 & 28.0 & 24.2 & 81.0 & 83.6 & 27.0 \\
Qwen3-VL-32B & 87.1 & 34.5 & 87.8 & 36.4 & 28.7 & 88.5 & 88.8 & 32.8 \\
Qwen3.8-27B & 72.7 & 72.8 & 79.6 & 68.3 & 56.3 & 79.7 & 77.5 & 65.2 \\
\midrule
MiMo-VL-7B-RL (Thinking) & 85.3 & 81.3 & 87.4 & 82.7 & 71.2 & 87.6 & 87.1 & 74.0 \\
Qwen3-VL-8B-Thinking & 85.7 & 59.0 & 86.0 & 59.8 & 51.0 & 86.5 & 87.0 & 48.2 \\
Qwen3-VL-32B-Thinking & 92.7 & 70.5 & 92.0 & 69.6 & 63.9 & 93.0 & 91.8 & 61.6 \\
Qwen3.8-27B (Thinking) & 94.2 & 84.2 & 94.5 & 84.3 & 79.9 & 95.0 & 94.7 & 79.1 \\
\bottomrule
\end{tabular}

\caption{\textbf{Accuracy per input and under evidence additions} (\%). $L$ \& $K$: insufficient $L$ and orderable $K$ both correct (1,100 pairs). $K$ (pairs): accuracy on $K$ over these pairs; $K\mid L$ correct: the same, given a correct $L$. $L$ \& $P$: both correct where their labels agree (1,156 pairs).}
\label{tab:evidence_response}
\end{table*}

\Cref{tab:evidence_response} gives accuracy per input and under the two additions. $K\mid L$ correct is the accuracy on $K$ among the pairs where $L$ was correctly answered UNCLEAR; it is at most five points below the accuracy on $K$ over all such pairs ($K$ (pairs)) for fifteen of the sixteen configurations and higher for ten; the exception, Keye-VL-1.5-8B (33.7\% vs.\ 49.0\%), often withholds on $K$ as well, as on other orderable inputs. Models that withhold on $L$ thus use the critical frames once they are added. Accordingly, most joint failures on $L$ and $K$ (63--100\%) come from committing on $L$ for every configuration except Gemma~4 (47.3\%), which also withholds on many orderable inputs. We also compare answer changes after a peripheral addition with changes under the A/B swap of the same correct $L$ input, which alters only the label assignment. For fourteen configurations, peripheral frames change correct answers no more often than the swap does; only Gemma~4 (+5.8 points, 95\% interval 3.1--8.5) and Qwen3-VL-32B-Thinking (+3.7, 0.6--6.6) show small excesses, so we do not read peripheral changes as distraction by irrelevant evidence. Accordingly, accuracy on $P$ stays within five points of that on $L$ for fifteen of the sixteen configurations (Gemma~4: 7.4 points lower).

\subsection{Prompt Strategies}\label{app:strategies}
\begin{figure*}[!t]
\centering
\includegraphics[width=\textwidth]{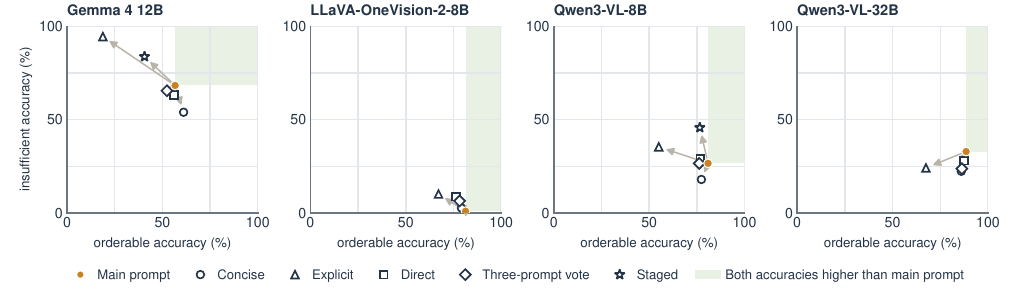}
\caption{\textbf{No prompt variant improves ordering and withholding together.} Accuracy on orderable and on insufficient inputs under the main prompt (filled) and each alternative (open), with arrows from the main prompt; no alternative falls in the shaded region, where both accuracies exceed those of the main prompt. Staged checking was run for Qwen3-VL-8B and Gemma~4.}
\label{fig:prompt_tradeoff}
\end{figure*}

\begin{table*}[!t]
\centering\scriptsize\setlength{\tabcolsep}{1.6pt}
\begin{tabular}{lrrrrrrr}
\toprule
Metric & Qwen3-VL 8B$\rightarrow$32B & InternVL3.5 8B$\rightarrow$14B & MiMo-VL SFT$\rightarrow$RL & MiMo-VL-RL Think. & Qwen3-VL-8B Think. & Qwen3-VL-32B Think. & Qwen3.8-27B Think. \\
\midrule
Overall & 7.0 [5.1, 8.9] & 0.8 [$-$0.7, 2.5] & 10.0 [8.5, 11.5] & 29.7 [27.4, 32.0] & 18.2 [16.0, 20.4] & 19.7 [17.7, 21.8] & 15.9 [13.8, 18.2] \\
Orderable & 7.7 [5.4, 10.1] & 2.7 [0.3, 5.2] & 5.9 [4.0, 7.7] & 19.3 [16.9, 21.7] & 5.8 [3.4, 8.4] & 4.7 [2.8, 6.6] & 18.0 [15.7, 20.3] \\
Insufficient & 6.3 [2.7, 9.8] & $-$1.3 [$-$3.5, 1.1] & 14.3 [11.9, 16.7] & 40.7 [36.9, 44.6] & 31.2 [27.5, 34.9] & 35.5 [32.0, 39.3] & 13.8 [10.1, 17.8] \\
$U_0$ & 7.8 [2.5, 13.1] & $-$4.2 [$-$9.0, 0.7] & 13.7 [9.2, 18.6] & 15.9 [10.6, 21.1] & 1.8 [$-$3.8, 7.5] & 5.2 [1.2, 9.5] & 0.5 [$-$3.0, 4.3] \\
$U_1$ & $-$16.2 [$-$24.1, $-$8.7] & 1.5 [$-$6.2, 10.1] & 12.9 [8.1, 18.1] & 30.2 [21.1, 39.0] & 27.1 [20.2, 34.3] & 45.6 [37.9, 53.7] & 30.8 [21.1, 40.1] \\
$U_2$ & 14.5 [10.1, 19.0] & $-$1.1 [$-$2.7, 0.5] & 15.2 [12.0, 18.4] & 56.1 [51.1, 61.0] & 46.1 [41.0, 51.2] & 45.3 [40.3, 50.6] & 13.1 [8.0, 18.7] \\
Family & 6.4 [3.5, 9.4] & 0.0 [$-$1.7, 1.7] & 4.8 [2.6, 7.0] & 40.8 [36.4, 45.1] & 15.4 [11.4, 19.4] & 25.2 [21.1, 29.4] & 34.5 [29.7, 39.4] \\
\bottomrule
\end{tabular}

\caption{\textbf{Paired checkpoint changes} (points, 95\% cluster-bootstrap intervals) on identical queries (for Family, on identical families). Thinking comparisons use first-pass outputs.}
\label{tab:paired}
\end{table*}

Four models answer every query under three further prompts and their majority vote (\cref{fig:prompts_variants}a--c), and two of them under staged checking, which appends the model's own occurrence answers to the main prompt (\cref{fig:prompts_variants}d). \Cref{fig:prompt_tradeoff} shows that no variant improves orderable and insufficient accuracy together: most trade one for the other, and some lower both. The explicit instruction raises Gemma~4's insufficient accuracy to 94.5\% but cuts its orderable accuracy to 18.8\%. Staged checking improves Qwen3-VL-8B by 7.1 overall points (95\% interval 5.7--8.5) but not Gemma~4 ($-0.7$, interval $-2.1$ to 0.7); it needs 9,264 calls per model, the 4,632 order queries plus 4,632 occurrence calls that are shared by the two assignments. The variants are not paraphrases of the main prompt: the explicit instruction also yields many invalid outputs for the Qwen3-VL models (614 and 426 of 4,632), and the vote, which combines different instructions rather than repeated samples, is invalid without a majority (21--154 queries).

\subsection{Checkpoints and Thinking Budget}\label{app:checkpoints}\label{app:cost}

\Cref{tab:paired} gives the paired changes of \cref{tab:comparisons} with intervals for all metrics. Except for the two Thinking switches, which keep the checkpoint and change only the prompt suffix (MiMo-VL-7B-RL) or the chat template's thinking setting (Qwen3.8-27B), the two checkpoints of a pair also differ in training data and recipe, so these are comparisons of released models rather than controlled interventions. Two comparisons with long-video refusal~\cite{virtuebench}: scaling Qwen3-VL from 8B to 32B raises accuracy on insufficient inputs by 6.3 points, in line with the higher refusal accuracy reported there for the larger model, although withholding falls on $U_1$; Thinking, by contrast, raises withholding here, whereas on long videos it lowered refusal accuracy for Qwen3-VL-32B, an exception to the higher refusal accuracy reported there for most reasoning variants. RL post-training of MiMo-VL coincides with a weaker preference for answer A (77.3\% to 59.4\% of outputs; \cref{tab:uncertainty_strata}).

Qwen3-VL-8B-Thinking truncates 250 of its first 4,096-token outputs. As a sensitivity analysis, each truncated query received one 8,192-token retry and the 81 still truncated one 16,384-token retry; retries depend only on truncation, and the last 37 truncations remain errors. Overall accuracy rises from 72.6\% to 74.5\% and 75.0\%, while the generated tokens grow from 4.25M to 5.91M and the summed generation time from 26.2 to 36.7 hours. The main results use the first pass. Qwen3-VL-32B-Thinking truncates 13 outputs and MiMo-VL-7B-RL with Thinking 43, which remain errors without retries; Qwen3.8-27B with Thinking truncates none.

\FloatBarrier
\section{Randomly Drawn Examples}\label{app:case}
\begin{figure*}[!t]
\centering
\includegraphics[width=\textwidth]{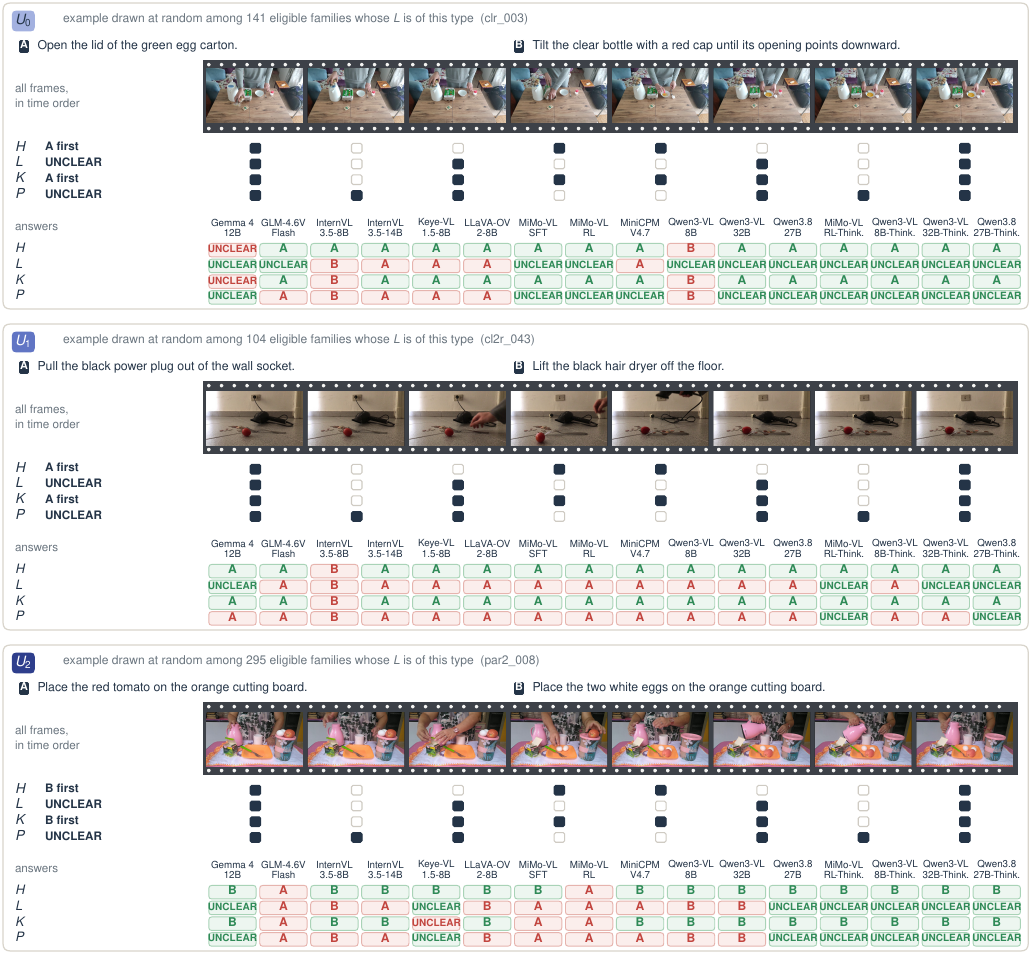}
\caption{\textbf{Randomly drawn examples.} One standard family per evidence type of $L$. Filled squares mark the frames of each input; answers use the original assignment (green: correct; red: incorrect).}
\label{fig:gallery}
\end{figure*}

\Cref{fig:gallery} shows one family per evidence type of $L$, drawn at random (seed 20261106) among the standard families of that type after excluding those shown elsewhere in the paper; model answers play no part in the selection. Each example gives all eight frames in temporal order, which frames each input contains, the reference answers, and the answers of all sixteen configurations. In the $U_1$ example, twelve configurations order the actions on $L$ although no frame of $L$ establishes that the hair dryer is lifted. In the $U_2$ example, nine configurations order the actions on $L$, and two answer A on all four inputs, although $H$ and $K$ establish that B comes first.

\section{Provenance, Release, and Data Use}\label{app:release}
Frozen protocols record the ordered image hashes, request texts, model revisions, and execution environments; reference labels are stored outside the requests. Raw outputs are kept separately from derived scores, analysis exports and paper tables are linked by hashes, and an independent audit recomputes parsing, coverage, input pairs, evidence types, diagnostic denominators, paired comparisons, and retry eligibility.

All frames come from the released source videos and are used under their licenses: CC BY 4.0 for Perception Test and HO-Cap, and Apache 2.0 for CaptainCook4D. Frames are at most resized: HO-Cap frames are 640$\times$480, and all other frames but those of one Perception Test family are 1,280 or 1,344 pixels on the longer side. In CaptainCook4D frames, the recipe display screen is masked before resizing, so on-screen recipe text is not part of the visual evidence. The annotations describe actions on objects and add no information about the people shown.

The event families, frames, labels, and prompts are released at \url{https://huggingface.co/datasets/xiaoqinfeng/EvOrderBench}, and the evaluation code with the answers of all evaluated configurations at \url{https://github.com/XqFeng-Josie/EvOrderBench}. The code is released under the MIT License and the annotations under CC BY 4.0; the frames keep the licenses of their source datasets.

\end{document}